\pdfoutput=1
\documentclass[11pt]{article}
\usepackage{EMNLP2023}

\usepackage{times}
\usepackage{latexsym}

\usepackage[T1]{fontenc}
\usepackage[utf8]{inputenc}

\usepackage{microtype}

\usepackage{inconsolata}

\usepackage{pifont}% http://ctan.org/pkg/pifont
\newcommand{\cmark}{\ding{51}}%

\usepackage{rotating}
\usepackage[many]{tcolorbox}
\usepackage{verbatim}

\usepackage{booktabs}
\usepackage{multirow}
\usepackage{multicol}

\usepackage{subcaption}

\usepackage[shortlabels]{enumitem}
\setlist[itemize]{leftmargin=*}
\setlist[enumerate]{leftmargin=*}
\usepackage[normalem]{ulem}

\title{Towards A Holistic Landscape of \\ Situated Theory of Mind in Large Language Models}

\author{
  Ziqiao Ma \hspace{30pt}  
  Jacob Sansom \hspace{30pt} Run Peng \hspace{30pt} 
  Joyce Chai \\
  Computer Science and Engineering Division, University of Michigan \\ 
  \texttt{\{marstin,jhsansom,roihn,chaijy\}@umich.edu} \\
}

\begin{document}

\maketitle

\begin{abstract}

Large Language Models (LLMs) have generated considerable interest and debate regarding their potential emergence of Theory of Mind (ToM).
Several recent inquiries reveal a lack of robust ToM in these models and pose a pressing demand to develop new benchmarks, as current ones primarily focus on different aspects of ToM and are prone to shortcuts and data leakage.
In this position paper, we seek to answer two road-blocking questions:
(1) How can we taxonomize a holistic landscape of machine ToM? 
(2) What is a more effective evaluation protocol for machine ToM?
Following psychological studies, we taxonomize machine ToM into 7 mental state categories and delineate existing benchmarks to identify under-explored aspects of ToM.
We argue for a holistic and situated evaluation of ToM to break ToM into individual components and treat LLMs as an agent who is physically situated in environments and socially situated in interactions with humans. 
Such situated evaluation provides a more comprehensive assessment of mental states and potentially mitigates the risk of shortcuts and data leakage.
We further present a pilot study in a grid world setup as a proof of concept.
We hope this position paper can facilitate future research to integrate ToM with LLMs and offer an intuitive means for researchers to better position their work in the landscape of ToM.

\end{abstract}

\section{Introduction}

The term \textit{theory of mind} (ToM, sometimes also referred to as \textit{mentalization} or \textit{mindreading}) was first introduced by~\citet{premack1978chimpanzee} as agents' ability to impute \textit{mental states} to themselves and others.
Many aspects of human cognition and social reasoning rely on ToM modeling of others' mental states~\citep{gopnik1992child,baron1997mindblindness,gunning2018machine}.
This is crucial for understanding and predicting others' actions~\citep{dennett1988precis}, planning over others' beliefs and next actions~\citep{ho2022planning}, and various forms of reasoning and decision-making~\citep{pereira2016integrating,rusch2020theory}.
Inspired by human ToM, AI researchers have made explicit and implicit efforts to develop a machine ToM for \textit{social intelligence}: AI agents that engage in social interactions with humans~\citep{kramer2012human,kennington2022understanding} and other agents~\citep{albrecht2018autonomous}.
A machine ToM enables an interactive paradigm of language processing~\citep{wang2023interactive}, enhancing agents' capacity for interactions~\citep{wang2021towards}, explainable decision-making~\citep{akula2022cx}, dialogue communication~\citep{qiu2022towards,takmaz2023speaking}, and collaborative task planning~\citep{bara2023towards}.

Machine ToM has received an increasing amount of attention, especially as the field is reshaped by \textit{large language models} (LLMs) such as ChatGPT~\citep{chatgpt2022openai} and GPT-4~\citep{openai2023gpt4}. 
This highlights an ongoing debate and discussion on whether a machine ToM has emerged in LLMs. 
While LLMs have demonstrated some capability of inferring communicative intentions, beliefs, and desires~\citep{andreas2022language,kosinski2023theory,bubeck2023sparks}, researchers also reported concerns regarding a lack of robust \textit{agency} in LLMs for complex social and belief reasoning tasks~\citep{sap2022neural,shapira2023clever} and in-context pragmatic communication~\citep{ruis2022large}. 
Emerged or not emerged, that remains a question (or may not even be the central question to ask). 
In our view, existing evaluation protocols do not fully resolve this debate. 
Most current benchmarks focus only on a few aspects of ToM, in the form of written stories. 
Echoing~\citet{trott2022large,aru2023mind,shapira2023clever}, many are also prone to data contamination, shortcuts, and spurious correlations.
Before extensive data collection for new ToM benchmarks, it is crucial to address two key questions: (1) How can we taxonomize a holistic landscape of machine ToM? (2) What is a more effective evaluation protocol for machine ToM?

To embrace the transformation brought by LLMs and explore their full potential in understanding and modeling ToM, this position paper calls for a holistic investigation that taxonomizes ToM using the \textit{Abilities in Theory of Mind Space} (ATOMS) framework~\cite{beaudoin2020systematic}.
After a review of existing benchmarks under this framework, we put forward a situated evaluation of ToM, one that treats LLMs as agents who are physically situated in environments and socially situated in interactions with humans. 
We hope this paper will offer an intuitive means to identify research priorities and to help gain a deeper understanding of, as well as to effectively utilize, LLMs in ToM modeling for AI agents in the future.

\section{Large Language Models as Theory of Mind Agents}
\label{sec::llm}

Since the advent of pre-trained language models, the research community has questioned whether they possess intrinsic mental states to represent the environment~\citep{li2021implicit,storks2021tiered,hase2023methods} and comprehend the mental states of others~\citep{sap2019social,zhang2021effectiveness} through the textual description (observation) of behavioral cues.
The relatively recent breakthroughs of LLMs have created many discussions and debates, primarily concerning the extent to which LLMs possess various capabilities required for a machine ToM. 
In this section, we first survey recent research presenting evidence and counter-evidence for the emergence of ToM in LLMs.
We conclude the discussion with the limitations of current evaluation protocols.

\subsection{Do Machine ToM Emerge in LLMs?}

\paragraph{Evidence for emergent ToM in LLMs.}
Prior to the rise of large language models, there has been growing evidence and acknowledgment of a narrow and limited sense of agency in smaller language models.
\citet{andreas2022language} argues that language models have the capacity to predict relations between agents' observations, mental states, actions, and utterances, as they infer approximate representations of beliefs, desires, and intentions of agents mentioned in the context.
These representations have a causal influence on the generated text, similar to an intentional agent's state influencing its communicative actions under a Belief-Desire-Intention (BDI) agent model~\citep{bratman1987intention}.
Amidst the excitement surrounding the release of GPT-4~\citep{openai2023gpt4}, researchers have searched for evidence of an emergent ToM in LLMs.
\citet{kosinski2023theory} presents 20 case studies each of the unexpected contents task~\citep{perner1987three} and the unexpected transfer (Sally-Anne) task~\citep{baron1985does}.
With direct comparisons to children's performance, the findings have been cited as potential evidence for a spontaneous emergence of ToM in LLMs.
\citet{bubeck2023sparks} present a similar behavioral study with 10 cases of belief, emotion, and intention understanding, concluding that GPT-4 has an advanced level of ToM after qualitative comparison with predecessors.
Other case studies have also shown aspects of machine ToM~\citep{li2023camel,holterman2023does}.

\vspace*{-5pt}
\paragraph{Limitations of ToM capabilities in LLMs.}
The above findings contradict the conclusions drawn in \citet{sap2022neural}'s earlier study, which shows a clear lack of ToM in GPT-3~\citep{brown2020language} on \textsc{SocialIQA}~\citep{sap2019social} and \textsc{ToMi}~\citep{le2019revisiting} benchmarks.
As a potential account, there has been criticism that the cognitive inquiries are anecdotal and inadequate for evaluating ToM in LLMs~\citep{marcus2023how,mitchell2023debate,shapira2023clever}.
Following the same evaluation protocol, \citet{ullman2023large} demonstrates that simple adversarial alternatives to \citet{kosinski2023theory} can fail LLMs.
To further understand if the most recent variants of LLMs possess a robust ToM, \citet{shapira2023clever} present a comprehensive evaluation over 6 tasks and 3 probing methods, showing that a robust machine ToM is absent even in GPT-4 and that LLMs are prone to shortcuts and spurious correlations. 
Based on the ongoing debate, it can be concluded that, while LLMs exhibit some level of sensitivity at understanding others' mental states, this capability is limited and falls short of achieving robust human-level ToM~\citep{trott2022large,shapira2023clever}.

\vspace*{-4pt}
\subsection{Roadblocks in ToM Evaluation in LLMs}

Given the pressing need for a robust machine ToM in LLMs and large-scale ToM benchmarks, researchers echo several difficulties in the evaluation protocol.
Presently, ToM benchmarks suffer from three primary issues summarized as follows.

\vspace*{-5pt}
\paragraph{Limited aspects of ToM.}
The evaluation of machine ToM lacks consistency in the literature due to the ambiguity surrounding the specific mental states being targeted. 
Existing benchmarks often focus on limited numbers of mental states, such as the \textit{intention}~\citep{yoshida2008game}, \textit{belief}~\citep{grant2017can}, \textit{emotion}~\citep{sap2019social}, and \textit{knowledge}~\citep{bara2021mindcraft} of another agent. 
While all of these are necessary building blocks of machine ToM, we echo \citet{shapira2023clever}'s concern that the ToM capability of LLMs may have been over-claimed based on evaluations from only a specific aspect of ToM.
To give a comprehensive assessment of a holistic machine ToM, a taxonomy is essential to enable researchers to effectively position their work with different focuses and priorities, which may be orthogonal to each other.

\vspace*{-5pt}
\paragraph{Data contamination.}
Data contamination refers to the lack of a verifiable train-test split that is typically established to test the ability of machine learning models to generalize~\citep{magar2022data}. 
LLMs typically learn from internet-scale data, potentially giving them access during training to the data used to test them~\citep{bubeck2023sparks,hagendorff2023machine}. 
For ToM evaluation specifically, the training corpora of LLMs may contain research papers detailing these psychological studies. 
Many past studies used identical or slightly altered language prompts to test LLMs, leading to potential contamination issues~\citep{ullman2023large}.
To critically evaluate the performance of LLMs on ToM tasks, researchers must have access to the datasets used to train them~\citep{dodge2021documenting}, which are unfortunately not available.

\vspace*{-5pt}
\paragraph{Shortcuts and spurious correlations.}
The availability of shortcuts and spurious features has triggered many concerns that a model may leverage them to perform highly on a benchmark without robustly acquiring the desired skill~\citep{sclar2023minding,ullman2023large,shapira2023clever}.
Recent findings suggest that LLMs tend to learn surface-level statistical correlations in compositional tasks, potentially leading to an illusion of systematic learning~\citep{dziri2023faith}.
In all likelihood, LLMs are capable of learning ToM shortcuts in a similar manner. 

\begin{figure*}[htbp]
    \centering
    \begin{minipage}[t]{0.7\textwidth}
        \centering
        % \hspace*{-15pt}
        \scalebox{1.0}{
        \includegraphics[width=1.05\linewidth]{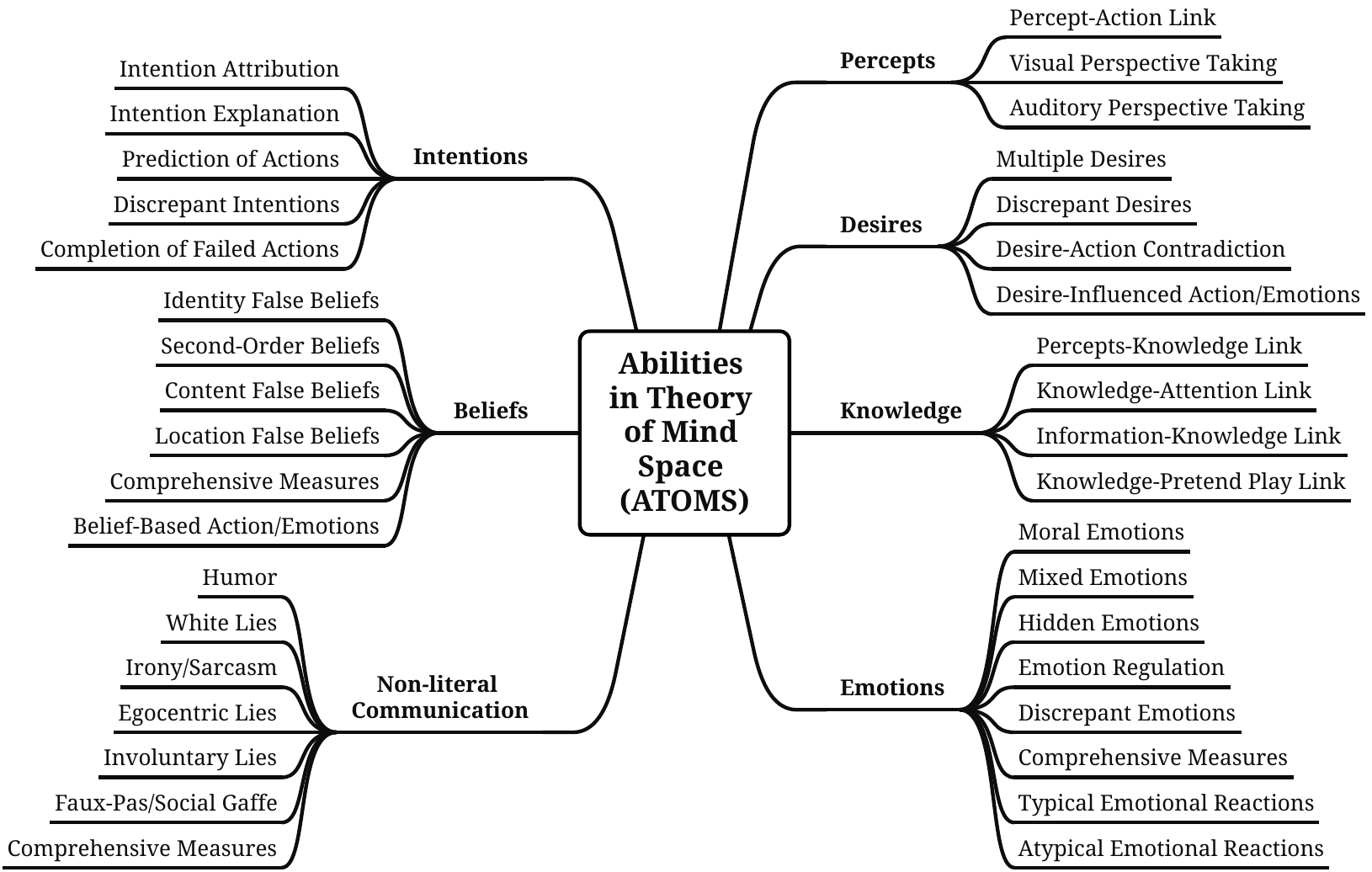}
        }
        \caption{The ATOMS framework of~\citet{beaudoin2020systematic}, which identified 7 categories of mental states through meta-analysis of ToM studies for children.}
        \label{fig::atoms}
    \end{minipage}\hfill
    \raisebox{1.0\height}{
    \begin{minipage}[t]{0.25\textwidth}    
        \begin{minipage}[t]{1\textwidth}
            \centering
            \scalebox{1.0}{
            \hspace*{-5pt}
            \includegraphics[width=1.05\linewidth]{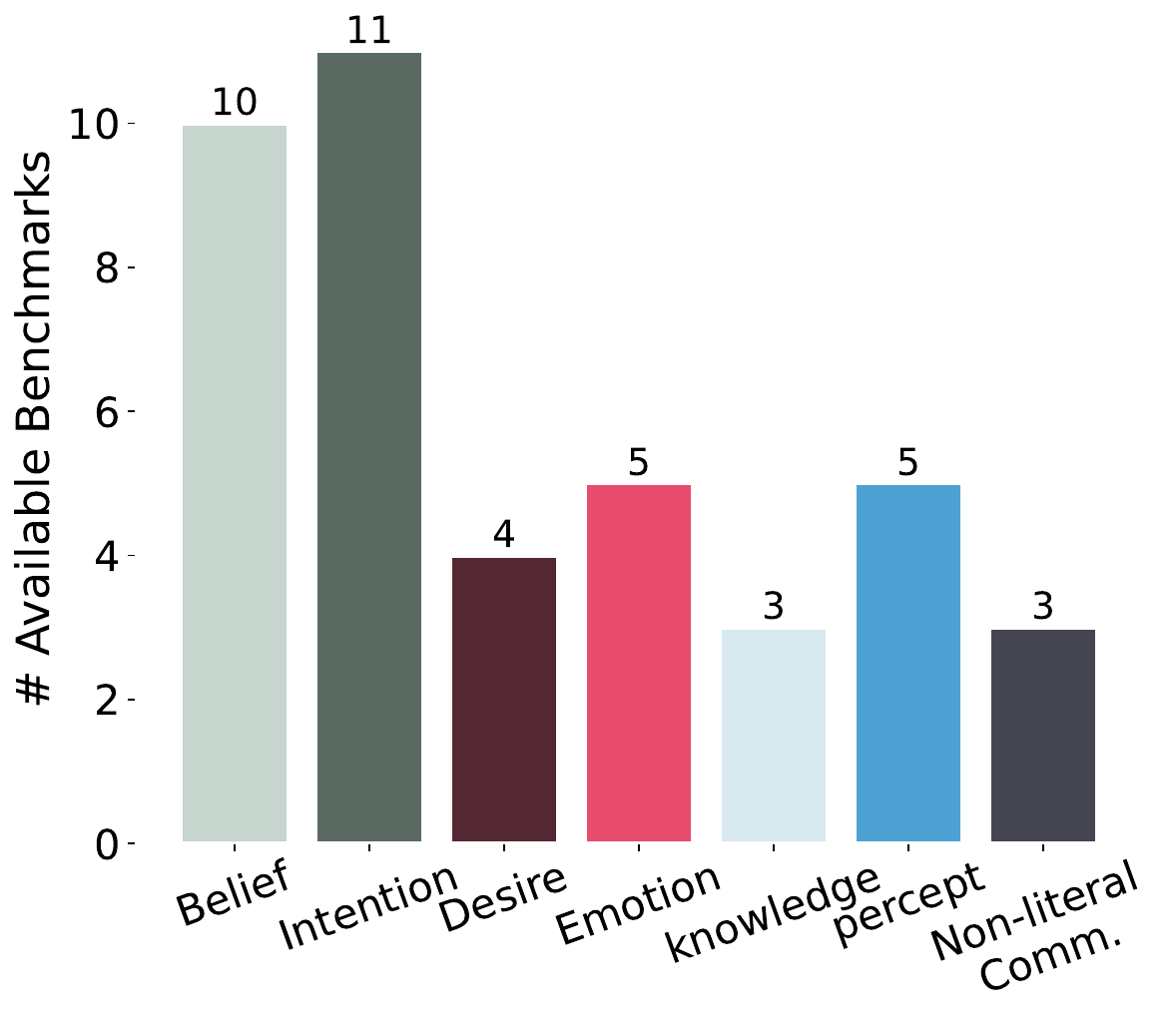
            }
            }
            \vspace*{-22pt}
            \caption{Number of available benchmarks for each mental state in ATOMS.}
            \label{fig:states}
        \end{minipage}\hfill
        \begin{minipage}[t]{1\textwidth}
            \centering
            \vspace{4pt}
            \scalebox{1.0}{
            \hspace*{-20pt}
            
            \includegraphics[width=1.2\linewidth]{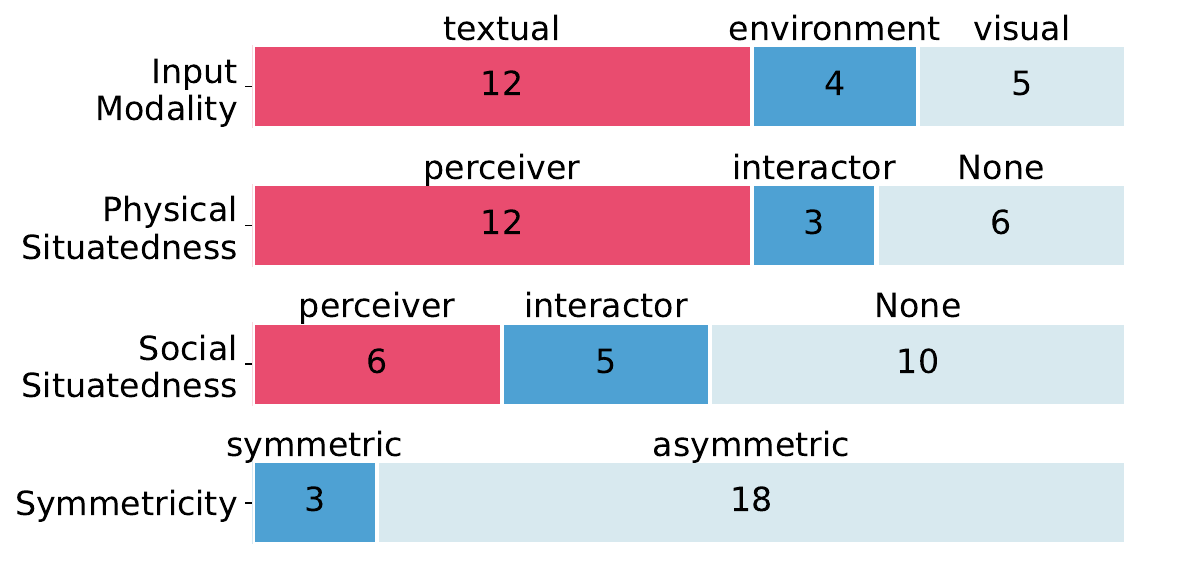}
            }
            \vspace*{-19pt}
            \caption{A comparison of benchmark settings.}
            \label{fig:setting}
        \end{minipage}
    \end{minipage}}
\end{figure*}
\begin{table*}[!ht]
\centering
\vspace{-8pt}
\scalebox{0.61}{
    \begin{tabular}{lccccccccccccccccl}
    \toprule
    \multirow{3}{*}{Benchmarks and Task Formulations}     & \multicolumn{3}{c}{Tested Agent}                                     & \multicolumn{4}{c}{Situatedness}                           & \multicolumn{9}{c}{ATOMS Mental States}                                                                                                                                                                            & \multirow{3}{*}{Sym.} \\ \cmidrule(r){2-4} \cmidrule(r){5-8} \cmidrule(r){9-17}
                                                     & \multirow{2}{*}{Task} & \multicolumn{2}{c}{Input Modality} & \multicolumn{2}{c}{Physical} & \multicolumn{2}{c}{Social} & \multicolumn{2}{c}{Belief} & \multicolumn{2}{c}{Intention} & \multirow{2}{*}{Des.} & \multirow{2}{*}{Emo.} & \multirow{2}{*}{Know.} & \multirow{2}{*}{Per.} & \multirow{2}{*}{NLC} &                           \\
                                                     &                              & Text      & Nonling.        & Per.      & Int.     & Per.     & Int.    & 1st & 2nd+ & Act.     & Com.    &                         &                          &                            &                          &                                    &                           \\ 
    \cmidrule(r){1-1} \cmidrule(r){2-4} \cmidrule(r){5-8} \cmidrule(r){9-17} \cmidrule(r){18-18}
    \rowcolor[HTML]{DAE8FC}
    \textsc{Epistemic Reasoning}~{\small\citep{cohen2021exploring}}   & Infer                        & T    &      -             &               &              &              &             & \cmark      & \cmark       &            &                  &                         &                          &                            &                          &                                    &                           \\
    \textsc{ToMi}~{\small\citep{nematzadeh2018evaluating}}            & QA                           & T    &     -              & \cmark        &              &              &             & \cmark      & \cmark       &            &                  &                         &                          &                            &                          &                                    &                           \\
    \rowcolor[HTML]{DAE8FC}
    \textsc{Hi-ToM}~{\small\citep{he2023hi}}            & QA                           & T    &      -             & \cmark        &              &              &             & \cmark      & \cmark       &            &                  &                         &                          &                            &                          &                                    &                           \\
    \textsc{MindGames}~{\small\citep{sileo2023mindgames}}             & Infer                        & T    &    -               & \cmark        &              &              &             & \cmark      & \cmark       &            &                  &                         &                          &                            & \cmark                   &                                    &                           \\
    \rowcolor[HTML]{DAE8FC}
    \textsc{Adv-CSFB}~{\small\citep{shapira2023clever}}               & QA                           & H    &     -              & \cmark        &              &              &             & \cmark      &              &            &                  &                         &                          &                            &                          &                                    &                           \\
    \textsc{ConvEntail}~{\small\citep{zhang2010towards}} & Infer                        & H    &       -            &               &              & \cmark       &             & \cmark      &              &            & \cmark           & \cmark                  &                          &                            &                          &                                    &                           \\
    \rowcolor[HTML]{DAE8FC}
    \textsc{SocialIQA}~{\small\citep{sap2019social}}                  & QA                           & H    &       -            &               &              & \cmark       &             &             &              & \cmark     &                  &                         & \cmark                   &                            &                          &                                    &                           \\
    \textsc{BeSt}~{\small\citep{tracey2022best}}                      & -                            & H    &       -            &               &              & \cmark       &             & \cmark      &              &            &                  &                         & \cmark                   &                            &                          & \cmark                             &                           \\
    \rowcolor[HTML]{DAE8FC}
    \textsc{FauxPas-EAI}~{\small\citep{shapira2023how}}               & QA                           & H,AI    &      -             &               &              & \cmark       &             & \cmark      &              &            &                  &                         &                          &                            &                          & \cmark                             &                           \\
    \textsc{COKE}~{\small\citep{wu2023coke}}                          & NLG                          & AI    &       -            &               &              & \cmark       & \cmark      &             &              & \cmark     &                  &                         & \cmark                   &                            &                          &                                    &                           \\
    \rowcolor[HTML]{DAE8FC}
    \textsc{ToM-in-AMC}~{\small\citep{yu2022few}}                     & Infer                        & H    &      -             & \cmark        &              & \cmark       &             &             &              & \cmark     & \cmark           &                         &                          &                            &                          &                                    &                           \\
    \textsc{G4C}~{\small\citep{zhou2023i}}                            & NLG                          & H,AI    &     -              & \cmark        &              & \cmark       & \cmark      &             &              & \cmark     & \cmark           &                         &                          &                            & \cmark                   &                                    &                           \\
    \rowcolor[HTML]{DAE8FC}
    \textsc{VisualBeliefs}~{\small\citep{eysenbach2016mistaken}}     & Infer                        &   -       & Cartoon            & \cmark        &              &              &             & \cmark      &              &            &                  &                         &                          &                            &                          & \cmark                             &                           \\
    \textsc{Triangle COPA}~{\small\citep{gordon2016commonsense}}      & QA                           & H    & Cartoon   & \cmark        &              & \cmark       &             &             &              & \cmark     &                  &                         & \cmark                   &                            &                          &                                    &                           \\
    \rowcolor[HTML]{DAE8FC}
    \textsc{MSED}~{\small\citep{jia2022beyond}}                       & Infer                        & H    & Images             & \cmark   & \multicolumn{1}{l}{} &          & \multicolumn{1}{l}{} &             &              &            &                  & \cmark                  & \cmark                   & \multicolumn{1}{l}{}       & \multicolumn{1}{l}{}     & \multicolumn{1}{l}{}               & \multicolumn{1}{l}{}      \\
    \textsc{BIB}~{\small\citep{gandhi2021baby}}                       & Infer                        &   -       & 2D Grid            & \cmark        &              &              &             &             &              & \cmark     &                  & \cmark                  &                          &                            &                          &                                    &                           \\
    \rowcolor[HTML]{DAE8FC}
    \textsc{AGENT}~{\small\citep{shu2021agent}}                       & Infer                        &    -      & 3D Sim.      & \cmark        &              &              &             &             &              & \cmark     &                  & \cmark                  &                          &                            & \cmark                   &                                    &                           \\
    \textsc{MToM}~{\small\citep{rabinowitz2018machine}}               & Infer                        &    -      & 2D Grid            & \cmark        &              &              &             & \cmark      &              & \cmark     &                  &                         &                          &                            &                          &                                    &                           \\
    \rowcolor[HTML]{DAE8FC}
    \textsc{SymmToM}~{\small\citep{sclar2022symmetric}}               & MARL                  &     -     & 2D Grid            & \cmark        & \cmark       & \cmark       & \cmark      &             &              &            &                  &                         &                          & \cmark                     &                          &                                    & \cmark                    \\
    \textsc{MindCraft}~{\small\citep{bara2021mindcraft}}              & Infer                        & H    & 3D Sim.      & \cmark        & \cmark       & \cmark       & \cmark      &             &              & \cmark     &                  &                         &                          & \cmark                     & \cmark                   &                                    & \cmark                    \\
    \rowcolor[HTML]{DAE8FC}
    \textsc{CPA}~{\small\citep{bara2023towards}}                      & Infer                        & H    & 3D Sim.      & \cmark        & \cmark       & \cmark       & \cmark      &             &              & \cmark     & \cmark           &                         &                          & \cmark                     & \cmark                   &                                    & \cmark                    \\ 
    \bottomrule
    \end{tabular}}
\vspace{-10pt}
\caption{A taxonomized review of existing benchmarks for machine ToM and their settings under ATOMS. We further break \textbf{beliefs} into first-order beliefs (\uline{1st}) and second-order beliefs or beyond (\uline{2nd+}); and break \textbf{intentions} into \uline{Act}ion intentions and \uline{Com}municative intentions. \textbf{Tasks} are divided into \uline{Infer}ence, \uline{Q}uestion \uline{A}nswering, \uline{N}atural \uline{L}anguage \uline{G}eneration, and \uline{M}ulti\uline{A}gent \uline{R}einforcement \uline{L}earning. \textbf{Input} modalities consist of \uline{Text} (\uline{H}uman, \uline{AI}, or \uline{T}emplate) and \uline{Nonling}uistic ones. The latter further breaks into \uline{Cartoon}, Natural \uline{Images}, \uline{2D Grid} World, and 3D \uline{Sim}ulation. The \textbf{Situatedness} is divided into None, Passive \uline{Per}ceiver, and Active \uline{Int}eractor. \textbf{Symmetricity} refers to whether the tested agent is co-situated and engaged in mutual interactions with other ToM agents. \vspace{-15pt}}
\label{tab:benchmarks}
\end{table*}

\section{Towards A Holistic Landscape of Machine Theory of Mind}
\label{sec::landscape}

\subsection{Abilities in Theory of Mind Space (ATOMS) Framework}

The evaluation of machine ToM lacks clarity and consistency across various literature, primarily due to the ambiguity surrounding the specific \textit{mental states} being targeted. 
This ambiguity is not unique to the field of AI but is rooted in the complicated cognitive underpinnings of ToM.
At the core of this ambiguity is the latent nature of \textit{mental states}, the subject has privileged access to them while others can only infer the existence of these mental states based on observable behaviors or expressions~\citep{dretske1979simple,blakemore2001perception,zaki2009unpacking}.
Thus, it is impossible to directly access and assess the mental states of a human, and ToM must be tested indirectly through humans' ability to understand the relationship between mental states and behaviors, especially by predicting how agents behave based on their mental states~\citep{swettenham1996can,phillips2002infants}.

While the exact definition of ToM remains a central debate, the AI community can benefit from looking at what psychologists have viewed as an initial step.
In this paper, we follow \citet{beaudoin2020systematic}'s taxonomy of ToM sub-domains, \textit{i.e.,} the Abilities in Theory of Mind Space (ATOMS).
As shown in Figure~\ref{fig::atoms}, the space consists of 7 categories of mental states, including \textit{beliefs}, \textit{intentions}, \textit{desires}, \textit{emotions}, \textit{knowledge}, \textit{percepts}, and \textit{non-literal communication}.
We selected this taxonomy because it was derived from a comprehensive meta-analysis of ToM studies.
The meta-analysis focused on young children aged 0-5 years at the early stage of cognitive development, such that the setups are simpler and more comparable, avoiding complicated physical and social engagements that cannot be trivially deployed on LLMs.

\vspace*{-5pt}
\paragraph{Beliefs.}
Beliefs are informational states that people judge to be true, usually decoupled from motivational states~\citep{dennett1995animals,eccles2002motivational}.
Beliefs, the most studied mental states in the field of ToM, are usually tested in the form of false belief tasks, including the unexpected contents test~\citep{perner1987three}, the unexpected transfer (Sally-Anne) Test~\citep{baron1985does}, the second-order false belief (Ice-cream Van) Test~\citep{perner1985john}. Researchers also studied their connection to actions and emotions~\citep{swettenham1996can}.

\vspace*{-5pt}
\paragraph{Intentions.}
Intentions are choices with commitment, usually associated with concrete actions towards a goal~\citep{cohen1990intention}.
As a critical component of ToM, \citet{kennington2022understanding} has called for a more explicit treatment of intentions.
Intentions have been extensively explored in psychology tests, e.g., behavioral re-enactment~\citep{meltzoff1995understanding}, action prediction~\citep{phillips2002infants}, intention explanation~\citep{smiley2001intention}, and intention attribution to abstract figures~\citep{castelli2006valley}.

\vspace*{-5pt}
\paragraph{Desires.}
Desires are motivational states that do not necessarily imply commitment, though they are usually emotionally charged and affect actions~\citep{malle2001distinction,kavanagh2005imaginary}.
Typical studies along this line include the Yummy-Yucky Task~\citep{repacholi1997early} for discrepant preferences from different individuals, the multiple desires within one individual~\citep{bennett1993children}, and the relationship between desires and emotions/actions~\citep{wellman1990simple,colonnesi2008precursors}.

\vspace*{-5pt}
\paragraph{Emotions.}
Emotions are mental states associated with an individual's feelings and affective experiences, which could impact beliefs and behaviors~\citep{frijda1986emotions,damasio2004emotions}.
Most ToM studies on emotions focus on typical~\citep{knafo2009empathy} and atypical~\citep{denham1986social} emotional reactions to situations.
Other studies also encompass affective perspective taking~\citep{borke1971interpersonal}, understanding hidden emotions~\citep{harris1986children}, and morally related emotions~\citep{pons2000test}.

\vspace*{-5pt}
\paragraph{Knowledge.}
Many controversies revolve around the definition of knowledge as justified true beliefs~\citep{gettier2000}.
In the context of AI, knowledge typically consists of information and organized representations of the world, which can be used to simplify understanding and address intricate reasoning and planning~\citep{schank2013scripts}.
ToM studies usually involve understanding the absence of knowledge~\citep{aronson1999preschoolers} as well as the connection between knowledge and perception~\citep{ruffman1989children} and attention~\citep{moll2006infants}.

\vspace*{-5pt}
\paragraph{Percepts.}
Humans are situated in the physical and social environments.
To enable AI agents to operate in the world and communicate with humans, the sensory and social aspects of perception are crucial in a machine ToM.
Along this line, psychological studies have investigated the perceptual perspective taking~\citep{masangkay1974early} and understanding the influence of limited perception on actions~\citep{hadwin1997does}.

\vspace*{-5pt}
\paragraph{Non-literal communications.}
Being able to understand non-literal and figurative communication helps humans to perform pragmatic inference and reason about hidden words behind their written meanings~\citep{giora2003our}.
Non-literal communication has been recognized as an advanced ToM capability, spanning a wide spectrum of humor and deceptions~\citep{happe1994advanced}, sarcasm~\citep{sullivan1995children}, and faux-pas (social gaffe) situations~\citep{baron1999recognition}.

\subsection{A Taxonomized Review of Benchmarks}

The ATOMS framework can serve as an intuitive reference for researchers to identify their research priorities and situate their work better in the landscape of literature.
We further take the initiative to provide a systematic review of existing benchmarks for machine ToM under the umbrella of ATOMS.\footnote{We maintain a repository for relevant literature at \url{https://github.com/Mars-tin/awesome-theory-of-mind}.}
Although there are independent research initiatives on certain ToM facets like intention classification, emotion modeling, and aspects of non-literal communications, we primarily focus on those that explicitly target ToM or inferences of latent mental states.
Besides the ToM dimensions in ATOMS, we further characterize the benchmarks on their task formulation, input modalities, physical and social situatedness, and symmetricity (whether the tested agent is co-situated and engaged in mutual interactions with other ToM agents).
We summarize our review in Table~\ref{tab:benchmarks} and discuss our observations and under-explored aspects of ToM evaluation.

\vspace*{-5pt}
\paragraph{Many aspects of ToM are under-explored.}
As shown in Figure~\ref{fig:states}, we notice an overwhelming research focus on the intention and belief aspects of machine ToM.
Several other aspects of ToM have not received enough attention.
While the field of NLP has thoroughly explored different facets of emotion and non-literal communication, e.g., in the context of dialogue systems, ToM has rarely been explicitly mentioned as motivation.
More connections and integrative efforts are clearly needed.

\vspace*{-5pt}
\paragraph{Lack of clear targeted mental states.}
Explicitly mentioning the Sally-Anne Test~\citep{baron1985does} as inspiration, \citet{grant2017can} developed the predecessor of \textsc{ToMi}~\citep{le2019revisiting}.
Similarly, \citet{nematzadeh2018evaluating} cited the Ice-cream Van Test~\citep{perner1985john} as motivation and the \textsc{FauxPas-EAI}~\citep{shapira2023how} benchmark followed the study of~\citet{baron1999recognition}.
While these benchmarks are cognitively grounded and target one particular aspect of ToM, the majority often incorporate multiple mental states without clear descriptions, which could make it challenging to measure the actual progress~\citep{raji2ai}.

\vspace*{-5pt}
\paragraph{Lack of situatedness in a physical and social environment.}
Figure~\ref{fig:setting} illustrates the configurations of benchmarks. 
Each bar in the chart represents a distinct benchmark characteristic, and each segment within the bar illustrates the proportion of benchmarks with one specific setting.
An immediate observation is a noticeable lack of benchmarks that encompass both physical and social environments, which highlights an existing research disparity in the field.
We notice that many existing benchmarks are story-based, which verbalize the agent's perception of the environment and the behaviors of other agents in the form of story episodes, usually with language templates.
The semantics of the environment are given by high-level events (e.g., Sally entered the kitchen).
Many aspects of physical and social situatedness are overlooked in these benchmarks, e.g., spatial relations, the task and motivation of agents, and their action trajectories.

\vspace*{-5pt}
\paragraph{Lack of engagement in environment.}
We point out that existing benchmarks primarily adopt a passive observer role to test language agents. 
Yet the crucial aspects of interaction and engagement between the agent and other entities involved have been overlooked.
Among all the benchmarks we reviewed, only three of them treat the tested model as an active agent, one that perceives the physical and social context, reasons about others' mental states, communicates with other agents, and interacts with the environment to complete pre-defined tasks~\citep{sclar2022symmetric,bara2021mindcraft,bara2023towards}.

\vspace*{-2pt}
\section{Towards A Situated Theory of Mind}
\label{sec::experiments}

\vspace*{-2pt}
\subsection{Why A Situated ToM?}

There have been concerns that cognitive inquiries are inadequate for gaining insight into understanding ToM for LLMs~\citep{mitchell2023debate,shapira2023clever}.
However, we believe that the primary problem lies in using story-based probing as proxies for psychological tests, which situate human subjects in specific physical or social environments and record their responses to various cues. 
We, therefore, call for a situated evaluation of ToM, in which the tested LLMs are treated like agents who are physically situated in environments and socially situated in interactions with others.

\vspace*{-5pt}
\paragraph{Situated evaluation covers more aspects of ToM.}
Although it is possible to frame the situations as narratives and cover all mental states using text-only benchmarks, certain aspects of ToM can only be effectively studied within specific physical or social environment~\citep{carruthers2015perceiving}. 
This is because humans have the ability to infer the mental states of others through various modalities such as visual perception, actions, attention (gazes or gestures), and speech~\citep{stack2022framework}. 
For instance, studying perceptual disparities can be challenging with text-only datasets, as they often reduce complex scenarios to rule-based manipulations over negations in the prompts~\citep{sileo2023mindgames}.
Benchmarks that are not situated also face challenges when it comes to implementing coordination between agents, e.g., aligning intentions towards joint actions~\citep{jain2019two} and pragmatic generation~\citep{zhu2021few,bao2022learning}.

\vspace*{-5pt}
\paragraph{Situated evaluation mitigates data contamination.}
A situated ToM evaluation can mitigate data contamination, as researchers can design scenarios in simulated settings that are unlikely to be part of the LLM's training data.
Carefully designed benchmarks can also incorporate seen and unseen environments to assess generalization to new tasks and new environments, fundamentally addressing the issue of data contamination~\citep{gandhi2021baby}.

\vspace*{-5pt}
\paragraph{Situated evaluation mitigates shortcuts.}
By employing situated evaluation, the risk of taking shortcuts can be mitigated. 
Many of the existing ToM benchmarks are either limited in scale or adopt text templates to verbalize a (few) predefined scenario(s) and prompt LLMs for answers, giving answers away from syntactic structures and positional information~\citep{le2019revisiting,sclar2023minding}.
In a situated setting, on the contrary, we rely on simulated environments to manipulate evaluation data at scale, so that the environment, the states, and the action traces in the environment can be randomized to avoid the statistical spurious correlations.
While situated evaluation can mitigate shortcuts, it does not eliminate the issue completely. 
For example, \citet{aru2023mind} have reported that shortcuts can emerge in grid world setups if the design is not careful enough and randomness is limited. 
We emphasize that careful design and consideration are still required to curate any ToM benchmark.

\begin{figure*}[!t]
    \centering
    \includegraphics[width=1.0\linewidth]{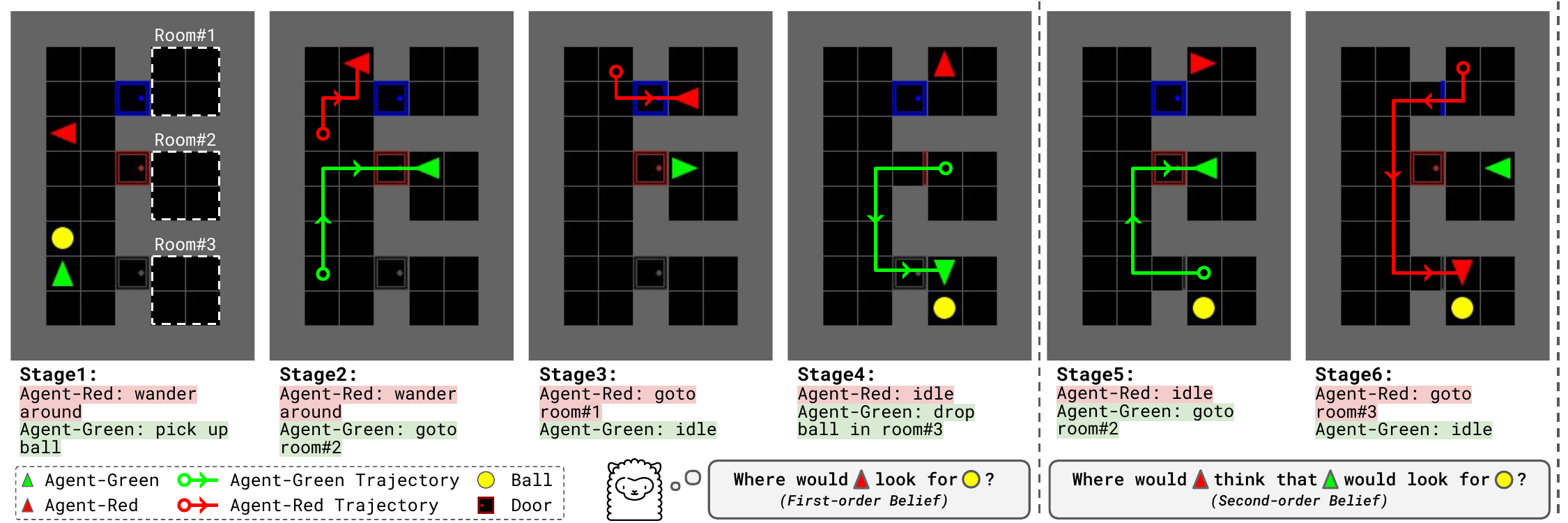}
    \vspace*{-20pt}
    \caption{An overview of the first and second order false belief task illustrated in a grid world setup. We simulate the unexpected transfer scenarios with two agents, and verbalize the environment and action traces to test if LLMs hold a correct understanding of the agents' false beliefs. \vspace*{-10pt}}
    \label{fig:task-belief}
\end{figure*}

\begin{figure*}[!t]
    \centering
    \includegraphics[width=1.0\linewidth]{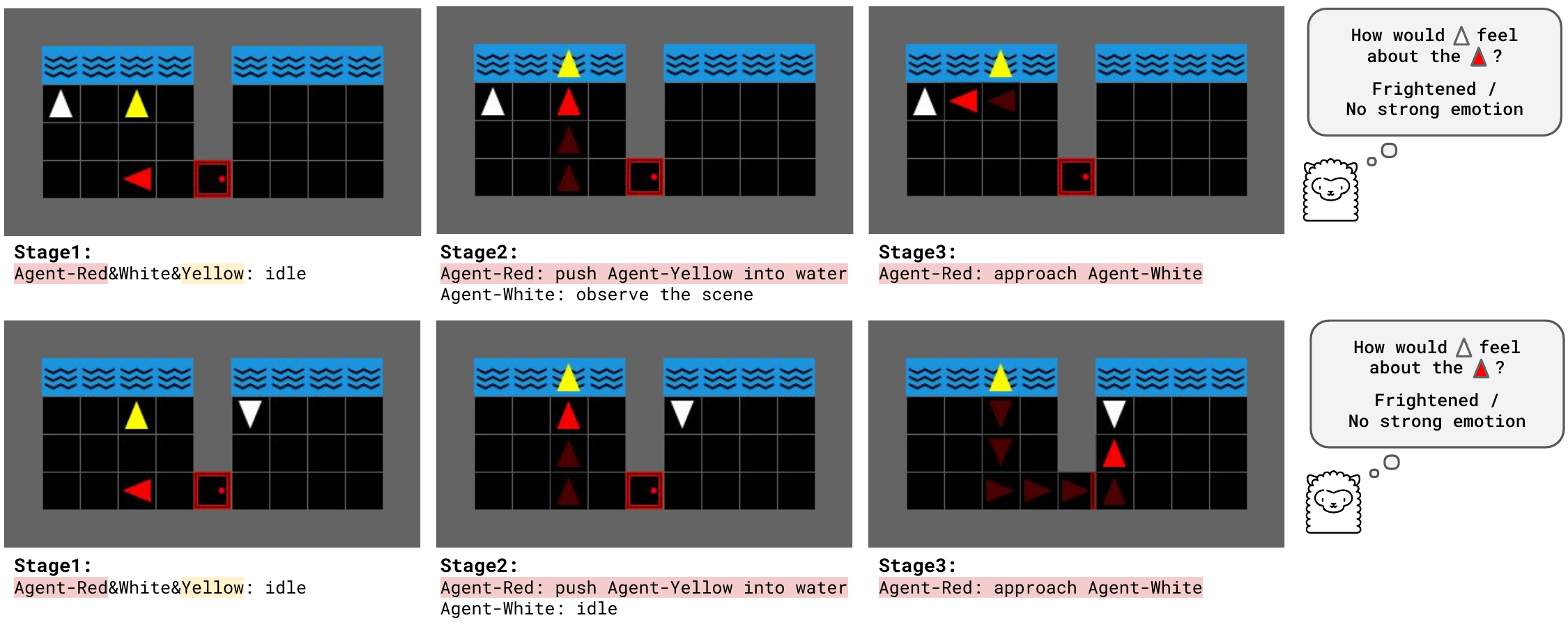}
    \vspace*{-20pt}
    \caption{An overview of the morally related emotional reaction tasks illustrated in a grid world setup. We simulate scenarios where an agent either directly witnesses or is ignorant of a morally related event, and verbalize the environment and action traces to test if LLMs hold a correct prediction of the agent's emotional reaction. \vspace*{-17pt}}
    \label{fig:task-emotion}
\end{figure*}

\vspace{-2pt}
\subsection{A Preliminary Exploration in Grid World}\label{sec::minigrid}

In this section, we present a proof-of-concept study on a situated evaluation of ToM on LLMs.
We choose to conduct our pilot study in MiniGrid~\citep{minigrid}, a simple and commonly used environment for ToM studies in the machine learning community~\citep{rabinowitz2018machine,sclar2022symmetric}.
Through basic grid world representation, we can create tasks to challenge LLMs to reason about many aspects of physical and social situatedness, e.g., spatial relations, partial observability, agent's action trajectories, and from there, their beliefs, intent, emotions, etc. 
This is in stark contrast to existing story-based ToM benchmarks, which only contain high-level event episodes.
We demonstrate that a diverse range of challenging ToM tests, covering all mental states from ATOMS, can be effectively created in a situated manner using a simple 2D grid world.

\vspace*{-5pt}
\paragraph{Environment and Task Setups}
We introduced 9 different ToM evaluation tasks for each mental state under ATOMS, and 1 reality-checking task to test LLMs' understanding of the world.
It is important to acknowledge that our experiment serves as a proof of concept and does not aim to cover the entire spectrum of machine ToM, as our case studies are far from being exhaustive or systematic. 
\vspace*{-15pt}
\begin{itemize}
    \setlength\itemsep{-0.25em}
    \item \textbf{Reality Check}: Given the sequence of actions, predict the closest object at the end of the trajectory. The task is designed to test LLMs' understanding of relocations in the grid world.
    \item \textbf{Short-term Intention}: Given an incomplete trajectory and a goal, predict the next action.
    \item \textbf{Long-term Intention}: Given an incomplete trajectory and a list of subgoals, predict the next subgoal that the agent is planning to achieve.
    \item \textbf{Desire}: Given a complete trajectory, predict if the agent demonstrates a preference for objects.
    \item \textbf{Percepts}: Given a complete trajectory, predict if the agent has a partial or full observation.
    \item \textbf{Belief}: The classic unexpected transfer task with possible first and second order false belief.
    \item \textbf{Non-literal Communication}: Given a trajectory and a statement from the agent, judge whether the agent is being deceptive.
    \item \textbf{Knowledge}: Given a trajectory, predict the object whose location is unknown to the agent.
    \item \textbf{Emotion}: The classic perception-emotion link test, where emotions are evoked in response to witnessing an emotionally stimulating situation.
\end{itemize}
\vspace*{-11pt}
We detail two case studies and leave examples of each task in Appendix~\ref{app::setting}.

\begin{figure*}[!t]
    \centering
    \includegraphics[width=\linewidth]{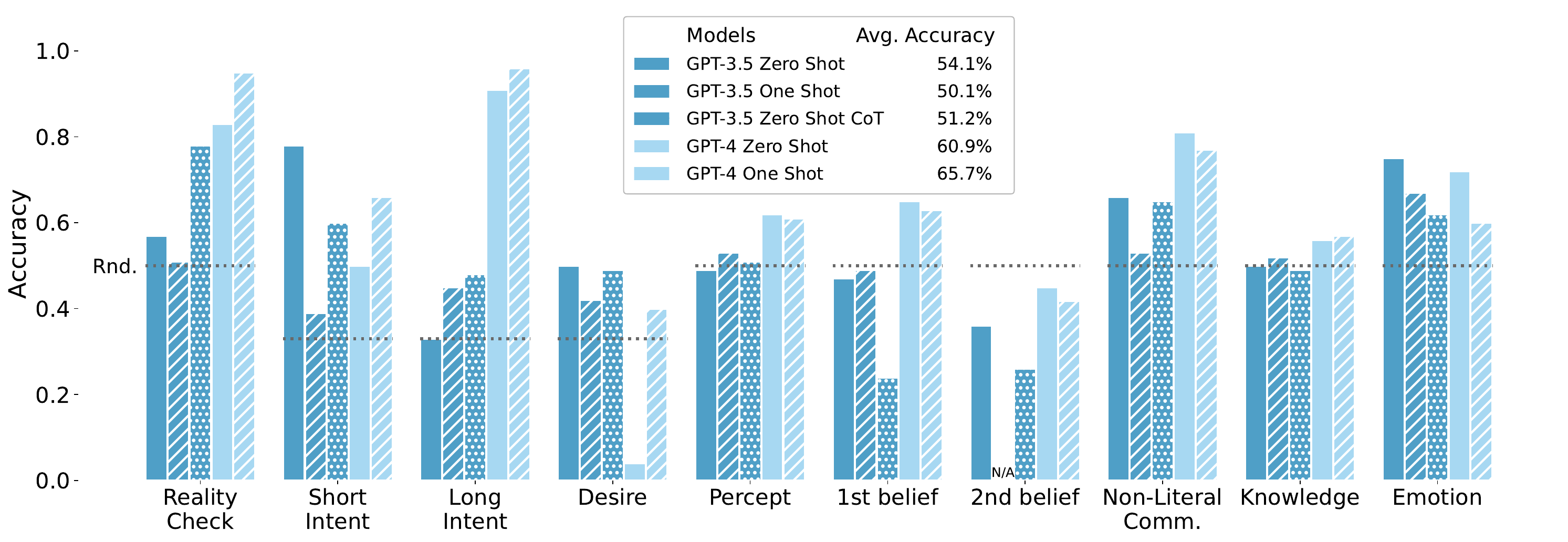}
    \vspace*{-21pt}
    \caption{The LLMs' performance across the 10 tasks is illustrated. Each bar shows how one LLM performed with a specific prompting method. Overall, the tasks are tough for all LLMs tested. The effectiveness of one-shot and CoT prompting is not consistent across the board. Some results are N/A as the prompt went out of the context window. \vspace*{-17pt}}
    \label{fig:results}
\end{figure*}

\vspace*{-5pt}
\paragraph{Case Study 1: Beliefs.}
Our belief experiments emulate the classic unexpected transfer tasks~\citep{baron1985does,perner1985john}.
As is shown in Figure~\ref{fig:task-belief}, we simulate this disparity of belief state and world state in MiniGrid.
The first-order belief task features a main room with three connected side rooms, two agents named \texttt{Red} and \texttt{Green}, and a ball.
Each instance of the belief experiment begins with \texttt{Green} placing the ball in \texttt{Room\#2} while \texttt{Red} watches.
\texttt{Red} then enters a separate \texttt{Room\#1} and shuts the door.
While \texttt{Red} is inside of this closed room, \texttt{Green} transfers the ball to \texttt{Room\#3}.
\texttt{Red} presumably holds a \textit{false belief} about the location of the ball, believing it is in \texttt{Room\#2} though it is now in \texttt{Room\#3}.
Similarly, we implement the second-order belief task to test an incorrect belief that one agent holds about the belief of another.
After \texttt{Green} has finished transferring the ball, it navigates to the room originally containing the ball and shuts the door.
\texttt{Red} then navigates to the room now containing the ball and sees the true location of the ball.
Still, \texttt{Green} presumably possesses a false belief about \texttt{Red}'s belief.
In both tasks, LLMs are queried with two versions of the world: a false one with the ball in the original room, and a true one with the ball in the third room (its actual location).
LLMs must correctly respond that the agents hold a false belief.

\vspace*{-5pt}
\paragraph{Case Study 2: Emotions.}
While the belief tasks highlight the importance of physical situatedness, we further demonstrate that social interactions can be simulated in the grid world.
As is shown in Figure~\ref{fig:task-emotion}, We design morally related events that stimulate emotions (e.g., fear, appreciation).
In this task, LLMs are queried to predict the emotional response of \texttt{Agent-White}, who either directly witnesses or is ignorant of this event.
LLMs must correctly respond that the agent holds an emotional reaction only if it observes the event.

\vspace*{-5pt}
\paragraph{Experiment Setups.}
For each task, we create 100 instances following a prompt template that consists of \texttt{[environment description]}, \texttt{[agent description]}, \texttt{[observability statement]}, \texttt{[task statement]}, \texttt{[actions sequences]}, \texttt{[QA]}.
We select GPT-4 (\texttt{gpt-4-0314}) and ChatGPT (\texttt{gpt-3.5-turbo-0613}) for evaluation on the 9 tasks.\footnote{We use the \texttt{ChatCompletion.create} function from \texttt{openai} package.}
Following prior work~\citep{hu2022fine,shapira2023clever}, we adopt MC-probing for LLMs that don't produce probabilities, which directly instructs LLMs to generate only the letter corresponding to the answer.
Besides zero-shot evaluation, we also explored one-shot learning and Chain-of-Thought (CoT) prompting~\citep{wei2022chain}.
More details are available in Appendix~\ref{sec::prompting}.

\vspace*{-5pt}
\paragraph{Results and Discussion.}
We observe that LLMs exhibit some level of sensitivity for some mental states.
Especially, GPT-4 scores up to 91\% zero-shot accuracy and 96\% one-shot accuracy in the long-term intention task. 
However, we also highlight the shortcomings of LLMs in some mental states of ATOMS to varying degrees, especially, in terms of predicting preferences, perception limitations, missing knowledge, and higher-order beliefs.
These findings align with previous research~\citep{sap2022neural,trott2022large,shapira2023clever}, further confirming that LLMs are not yet reliable and comprehensive ToM agents. 
From the reality-checking task, we observe that GPT-3.5 reaches 78\% accuracy with CoT prompting and GPT-4 significantly surpasses its predecessors with 83\% zero-shot accuracy and 95\% one-shot accuracy. 
Solving this reality check by no means implies that LLMs have a general perception ability of the real world, but that as a proof of concept, they demonstrate a certain (but still limited) level of situated awareness within the context of a basic abstract grid world.
This implies that researchers can begin utilizing them as powerful building blocks for situated agents in complex ToM tasks. 
We note that it is always possible to come up with more challenging reality-checking questions to expose the limitations of LLMs, or to provide more guided prompts to assist LLMs in successfully completing ToM tasks.
Undoubtedly, further research is required along this exciting yet challenging trajectory to advance ToM in LLMs and AI agents built upon LLMs.

\vspace*{-2pt}
\section{Discussions and Action Items}
\label{sec::discussion}

\vspace*{-2pt}
\subsection{The Scope of Machine Theory of Mind}

\paragraph{Be specific about the mental states studied.} 
Existing benchmarks often lack a clear target mental state, making it challenging to interpret the results and measure the actual progress. 
To mitigate the risk of overestimating LLMs' ToM capabilities, it is recommended that future benchmark developers provide specific details regarding the targeted mental state(s) they intend to assess.

\vspace*{-5pt}
\paragraph{Broaden the Scope of Machine ToM.} 
A breadth of mental states and their sub-domains have already been covered by AI benchmarks (Table~\ref{tab:benchmarks}).
We observed an overwhelming emphasis on the benchmarks and modeling of \textit{beliefs} and \textit{intentions}, while other aspects have received insufficient attention. 
Still, there are considerably many blank spaces in the landscape of machine ToM, especially for more complicated forms of knowledge, desires, perspective-tasking, and emotional experiences beyond typical social situations.

\vspace*{-2pt}
\subsection{Design New Theory of Mind Benchmarks}

\paragraph{Avoid shortcuts and spurious correlations.}
The evaluation of LLMs itself presents significant challenges, not only in the case of ToM. 
Existing benchmarks suffer from issues such as data leakage and spurious correlations.
Especially, shortcut solutions have been consistently reported in recent years~\citep{le2019revisiting,shapira2023clever,aru2023mind}.
We are in pressing need of new benchmarks with scalable sizes, high-quality human annotations, and privately held-out sets for evaluation.

\vspace*{-5pt}
\paragraph{Avoid unfair evaluations from prompting.}
Previous work has shown that CoT prompting can improve the performance of LLMs in ToM tasks~\citep{li2023camel,moghaddam2023boosting,shapira2023clever}. 
Various recent prompting mechanisms have also been developed to improve LLM's capability on ToM tasks~\citep{zhou2023far,leer2023violation}.
In the evaluation of LLMs' ToM capabilities, we recommend the careful documentation of prompts used and the avoidance of implicit human guidance to ensure a fair comparison.

\vspace*{-5pt}
\paragraph{Move on to a situated ToM.}
We call for a situated evaluation of ToM, in which the tested LLMs are treated like agents who are physically situated in environments and socially situated in interactions with others. 
A situated setup covers a wider range of ToM aspects. 
With carefully designed benchmarks with diverse environments and unseen test sets, a situated setup can help address data contamination issues and assess generalization to new tasks and environments.
Furthermore, a situated setup allows for more complicated evaluation protocols than simple inference and QA tasks.

\vspace*{-5pt}
\paragraph{Consider a mutual and symmetric ToM.}
ToM is symmetric and mutual in nature, as it originally imputes the mental states of self and others.
Prior research is largely limited to passive observer roles~\citep{grant2017can,nematzadeh2018evaluating,le2019revisiting,rabinowitz2018machine} or speaker in a speaker-listener relationship~\citep{zhu2021fewshot,zhou2023i}. 
We encourage more studies on how humans and agents build and maintain common ground with a human ToM and a machine ToM through situated communication~\citep{bara2021mindcraft,sclar2022symmetric}.
Besides, more research is needed to understand if LLMs possess early forms of intrinsic mental states given observation cues of the world.
While we need to develop machines that impute the mental states of humans, humans should also develop a theory of AI’s mind (ToAIM)~\citep{chandrasekaran2017takes} by understanding the strengths, weaknesses, beliefs, and quirks of these black box language models.

\subsection{Neural Language Acquisition and ToM}

Both psychological studies~\citep{bloom2002children,tomasello2005constructing} and computational simulations~\citep{liu2023computational} have demonstrated the effectiveness of ToM, especially intention, in language acquisition. 
Instead of concentrating on eliciting ToM in LLMs, we should contemplate whether certain ToM elements should be inherently present in LLMs or perhaps introduced alongside language pretraining.
More research is needed to understand the connection between neural word acquisition and ToM development in machines.

\vspace*{-2pt}
\section{Conclusion}
\label{sec::conclusion}
\vspace*{-2pt}

In this position paper, we survey and summarize the ongoing debate regarding the presence of a machine ToM within LLMs, and identify the inadequate evaluation protocols as the roadblock.
Many benchmarks focus only on a few aspects of ToM, and are prone to shortcuts.
To mediate this issue, we follow the ATOMS framework to offer a holistic review of existing benchmarks and identify under-explored aspects of ToM.
We further call for a situated evaluation of ToM, one that is physically situated in environments and socially situated in interactions with humans.
We hope this work can facilitate future research towards LLMs as ToM agents, and offer an intuitive means for researchers to position their work in the landscape of ToM.

\section*{Ethical Statement}

The dataset created in this study includes instances that are synthetically generated from planners and RL algorithms, as well as ones created by humans.
Human subjects research is approved by the University of Michigan Health Sciences and Behavioral Sciences Institutional Review Board (IRB-HSBS) under eResearch ID \texttt{HUM00234647}.
The text generated by LLMs could potentially contain harmful, toxic, or offensive content.
The authors have ensured that the data does not contain personally identifiable information or offensive content.

\section*{Limitations}

Our current benchmark only covers 100 instances for each task, adding up to only 1000 instances.
Our experiment serves as a proof of concept and does not aim to cover the entire spectrum of machine ToM, as our case studies are far from being exhaustive or systematic. 
In the future, we plan to create a more systematic benchmark with a larger scale and various forms of evaluation.
Additionally, it is worth noting that the ATOMS framework is derived from human ToM studies conducted with children under the age of 5. Consequently, this framework primarily focuses on the early developmental stages of ToM, capturing the naive and potentially rudimentary aspects of ToM.
For more advanced ToM capability, we point to some recent frameworks proposed by~\citet{osterhaus2022looking} and~\citet{stack2022framework}.
\section*{Acknowledgements}

This work was supported in part by 
NSF IIS-1949634, NSF SES-2128623, and by the Automotive Research Center at the University of Michigan.
Without implying any agreement with the contents as presented in this work, the authors extend their appreciation to Susan Gelman for her valuable feedback.
The authors would like to thank all anonymous reviewers for their valuable feedback.

\bibliographystyle{acl_natbib}
\bibliography{references}

\begin{thebibliography}{124}
\expandafter\ifx\csname natexlab\endcsname\relax\def\natexlab#1{#1}\fi

\bibitem[{Akula et~al.(2022)Akula, Wang, Liu, Saba-Sadiya, Lu, Todorovic, Chai, and Zhu}]{akula2022cx}
Arjun~R Akula, Keze Wang, Changsong Liu, Sari Saba-Sadiya, Hongjing Lu, Sinisa Todorovic, Joyce Chai, and Song-Chun Zhu. 2022.
\newblock Cx-tom: Counterfactual explanations with theory-of-mind for enhancing human trust in image recognition models.
\newblock \emph{Iscience}, 25(1):103581.

\bibitem[{Albrecht and Stone(2018)}]{albrecht2018autonomous}
Stefano~V Albrecht and Peter Stone. 2018.
\newblock Autonomous agents modelling other agents: A comprehensive survey and open problems.
\newblock \emph{Artificial Intelligence}, 258:66--95.

\bibitem[{Andreas(2022)}]{andreas2022language}
Jacob Andreas. 2022.
\newblock Language models as agent models.
\newblock In \emph{Findings of the Association for Computational Linguistics: EMNLP 2022}, pages 5769--5779, Abu Dhabi, United Arab Emirates. Association for Computational Linguistics.

\bibitem[{Aronson and Golomb(1999)}]{aronson1999preschoolers}
James~N Aronson and Claire Golomb. 1999.
\newblock Preschoolers' understanding of pretense and presumption of congruity between action and representation.
\newblock \emph{Developmental Psychology}, 35(6):1414.

\bibitem[{Aru et~al.(2023)Aru, Labash, Corcoll, and Vicente}]{aru2023mind}
Jaan Aru, Aqeel Labash, Oriol Corcoll, and Raul Vicente. 2023.
\newblock Mind the gap: Challenges of deep learning approaches to theory of mind.
\newblock \emph{Artificial Intelligence Review}, pages 1--16.

\bibitem[{Bao et~al.(2022)Bao, Ghosh, and Chai}]{bao2022learning}
Yuwei Bao, Sayan Ghosh, and Joyce Chai. 2022.
\newblock Learning to mediate disparities towards pragmatic communication.
\newblock In \emph{Proceedings of the 60th Annual Meeting of the Association for Computational Linguistics (ACL)}.

\bibitem[{Bara et~al.(2023)Bara, Ma, Yu, Shah, and Chai}]{bara2023towards}
Cristian-Paul Bara, Ziqiao Ma, Yingzhuo Yu, Julie Shah, and Joyce Chai. 2023.
\newblock Towards collaborative plan acquisition through theory of mind modeling in situated dialogue.
\newblock In \emph{Proceedings of the Thirty-Second International Joint Conference on Artificial Intelligence, {IJCAI-23}}, pages 2958--2966. International Joint Conferences on Artificial Intelligence Organization.
\newblock Main Track.

\bibitem[{Bara et~al.(2021)Bara, Sky, and Chai}]{bara2021mindcraft}
Cristian-Paul Bara, CH-Wang Sky, and Joyce Chai. 2021.
\newblock Mindcraft: Theory of mind modeling for situated dialogue in collaborative tasks.
\newblock In \emph{Proceedings of the 2021 Conference on Empirical Methods in Natural Language Processing}, pages 1112--1125.

\bibitem[{Baron-Cohen(1997)}]{baron1997mindblindness}
Simon Baron-Cohen. 1997.
\newblock \emph{Mindblindness: An essay on autism and theory of mind}.
\newblock MIT press.

\bibitem[{Baron-Cohen et~al.(1985)Baron-Cohen, Leslie, and Frith}]{baron1985does}
Simon Baron-Cohen, Alan~M Leslie, and Uta Frith. 1985.
\newblock Does the autistic child have a “theory of mind”?
\newblock \emph{Cognition}, 21(1):37--46.

\bibitem[{Baron-Cohen et~al.(1999)Baron-Cohen, O'riordan, Stone, Jones, and Plaisted}]{baron1999recognition}
Simon Baron-Cohen, Michelle O'riordan, Valerie Stone, Rosie Jones, and Kate Plaisted. 1999.
\newblock Recognition of faux pas by normally developing children and children with asperger syndrome or high-functioning autism.
\newblock \emph{Journal of autism and developmental disorders}, 29:407--418.

\bibitem[{Beaudoin et~al.(2020)Beaudoin, Leblanc, Gagner, and Beauchamp}]{beaudoin2020systematic}
Cindy Beaudoin, {\'E}lizabel Leblanc, Charlotte Gagner, and Miriam~H Beauchamp. 2020.
\newblock Systematic review and inventory of theory of mind measures for young children.
\newblock \emph{Frontiers in psychology}, 10:2905.

\bibitem[{Bennett and Galpert(1993)}]{bennett1993children}
Mark Bennett and Linda Galpert. 1993.
\newblock Children's understanding of multiple desires.
\newblock \emph{International Journal of Behavioral Development}, 16(1):15--33.

\bibitem[{Blakemore and Decety(2001)}]{blakemore2001perception}
Sarah-Jayne Blakemore and Jean Decety. 2001.
\newblock From the perception of action to the understanding of intention.
\newblock \emph{Nature reviews neuroscience}, 2(8):561--567.

\bibitem[{Bloom(2002)}]{bloom2002children}
Paul Bloom. 2002.
\newblock \emph{How children learn the meanings of words}.
\newblock MIT press.

\bibitem[{Borke(1971)}]{borke1971interpersonal}
Helene Borke. 1971.
\newblock Interpersonal perception of young children: Egocentrism or empathy?
\newblock \emph{Developmental psychology}, 5(2):263.

\bibitem[{Bratman(1987)}]{bratman1987intention}
Michael Bratman. 1987.
\newblock \emph{Intention, plans, and practical reason}.
\newblock The University of Chicago Press.

\bibitem[{Brown et~al.(2020)Brown, Mann, Ryder, Subbiah, Kaplan, Dhariwal, Neelakantan, Shyam, Sastry, Askell et~al.}]{brown2020language}
Tom Brown, Benjamin Mann, Nick Ryder, Melanie Subbiah, Jared~D Kaplan, Prafulla Dhariwal, Arvind Neelakantan, Pranav Shyam, Girish Sastry, Amanda Askell, et~al. 2020.
\newblock Language models are few-shot learners.
\newblock \emph{Advances in neural information processing systems}, 33:1877--1901.

\bibitem[{Bubeck et~al.(2023)Bubeck, Chandrasekaran, Eldan, Gehrke, Horvitz, Kamar, Lee, Lee, Li, Lundberg et~al.}]{bubeck2023sparks}
S{\'e}bastien Bubeck, Varun Chandrasekaran, Ronen Eldan, Johannes Gehrke, Eric Horvitz, Ece Kamar, Peter Lee, Yin~Tat Lee, Yuanzhi Li, Scott Lundberg, et~al. 2023.
\newblock Sparks of artificial general intelligence: Early experiments with gpt-4.
\newblock \emph{arXiv preprint arXiv:2303.12712}.

\bibitem[{Carruthers(2015)}]{carruthers2015perceiving}
Peter Carruthers. 2015.
\newblock Perceiving mental states.
\newblock \emph{Consciousness and cognition}, 36:498--507.

\bibitem[{Castelli(2006)}]{castelli2006valley}
Fulvia Castelli. 2006.
\newblock The valley task: Understanding intention from goal-directed motion in typical development and autism.
\newblock \emph{British journal of developmental psychology}, 24(4):655--668.

\bibitem[{Chandrasekaran et~al.(2017)Chandrasekaran, Yadav, Chattopadhyay, Prabhu, and Parikh}]{chandrasekaran2017takes}
Arjun Chandrasekaran, Deshraj Yadav, Prithvijit Chattopadhyay, Viraj Prabhu, and Devi Parikh. 2017.
\newblock It takes two to tango: Towards theory of ai's mind.
\newblock \emph{arXiv preprint arXiv:1704.00717}.

\bibitem[{Chevalier-Boisvert et~al.(2018)Chevalier-Boisvert, Willems, and Pal}]{minigrid}
Maxime Chevalier-Boisvert, Lucas Willems, and Suman Pal. 2018.
\newblock \href {https://github.com/Farama-Foundation/Minigrid} {Minimalistic gridworld environment for gymnasium}.

\bibitem[{Cohen(2021)}]{cohen2021exploring}
Michael Cohen. 2021.
\newblock Exploring roberta's theory of mind through textual entailment.

\bibitem[{Cohen and Levesque(1990)}]{cohen1990intention}
Philip~R Cohen and Hector~J Levesque. 1990.
\newblock Intention is choice with commitment.
\newblock \emph{Artificial intelligence}, 42(2-3):213--261.

\bibitem[{Colonnesi et~al.(2008)Colonnesi, Rieffe, Koops, and Perucchini}]{colonnesi2008precursors}
Cristina Colonnesi, Carolien Rieffe, Willem Koops, and Paola Perucchini. 2008.
\newblock Precursors of a theory of mind: A longitudinal study.
\newblock \emph{British Journal of Developmental Psychology}, 26(4):561--577.

\bibitem[{Damasio(2004)}]{damasio2004emotions}
Antonio~R Damasio. 2004.
\newblock Emotions and feelings.
\newblock In \emph{Feelings and emotions: The Amsterdam symposium}, volume~5, pages 49--57. Cambridge University Press Cambridge.

\bibitem[{Denham(1986)}]{denham1986social}
Susanne~A Denham. 1986.
\newblock Social cognition, prosocial behavior, and emotion in preschoolers: Contextual validation.
\newblock \emph{Child development}, pages 194--201.

\bibitem[{Dennett(1995)}]{dennett1995animals}
Daniel Dennett. 1995.
\newblock Do animals have beliefs.
\newblock \emph{Comparative approaches to cognitive science}, 111.

\bibitem[{Dennett(1988)}]{dennett1988precis}
Daniel~C Dennett. 1988.
\newblock Pr{\'e}cis of the intentional stance.
\newblock \emph{Behavioral and brain sciences}, 11(3):495--505.

\bibitem[{Dodge et~al.(2021)Dodge, Sap, Marasovi{\'c}, Agnew, Ilharco, Groeneveld, Mitchell, and Gardner}]{dodge2021documenting}
Jesse Dodge, Maarten Sap, Ana Marasovi{\'c}, William Agnew, Gabriel Ilharco, Dirk Groeneveld, Margaret Mitchell, and Matt Gardner. 2021.
\newblock Documenting large webtext corpora: A case study on the colossal clean crawled corpus.
\newblock In \emph{Proceedings of the 2021 Conference on Empirical Methods in Natural Language Processing}, pages 1286--1305.

\bibitem[{Dretske(1979)}]{dretske1979simple}
Fred~I Dretske. 1979.
\newblock Simple seeing.
\newblock In \emph{Body, mind, and method}, pages 1--15. Springer.

\bibitem[{Dziri et~al.(2023)Dziri, Lu, Sclar, Li, Jian, Lin, West, Bhagavatula, Bras, Hwang et~al.}]{dziri2023faith}
Nouha Dziri, Ximing Lu, Melanie Sclar, Xiang~Lorraine Li, Liwei Jian, Bill~Yuchen Lin, Peter West, Chandra Bhagavatula, Ronan~Le Bras, Jena~D Hwang, et~al. 2023.
\newblock Faith and fate: Limits of transformers on compositionality.
\newblock \emph{arXiv preprint arXiv:2305.18654}.

\bibitem[{Eccles and Wigfield(2002)}]{eccles2002motivational}
Jacquelynne~S Eccles and Allan Wigfield. 2002.
\newblock Motivational beliefs, values, and goals.
\newblock \emph{Annual review of psychology}, 53(1):109--132.

\bibitem[{Eysenbach et~al.(2016)Eysenbach, Vondrick, and Torralba}]{eysenbach2016mistaken}
Benjamin Eysenbach, Carl Vondrick, and Antonio Torralba. 2016.
\newblock Who is mistaken?
\newblock \emph{arXiv preprint arXiv:1612.01175}.

\bibitem[{Frijda et~al.(1986)}]{frijda1986emotions}
Nico~H Frijda et~al. 1986.
\newblock \emph{The emotions}.
\newblock Cambridge University Press.

\bibitem[{Gandhi et~al.(2021)Gandhi, Stojnic, Lake, and Dillon}]{gandhi2021baby}
Kanishk Gandhi, Gala Stojnic, Brenden~M Lake, and Moira~R Dillon. 2021.
\newblock Baby intuitions benchmark (bib): Discerning the goals, preferences, and actions of others.
\newblock \emph{Advances in Neural Information Processing Systems}, 34:9963--9976.

\bibitem[{Gettier(2000)}]{gettier2000}
Edmund~L Gettier. 2000.
\newblock Is justified true belief knowledge?
\newblock \emph{Causal Theories of Mind}, page 135.

\bibitem[{Giora(2003)}]{giora2003our}
Rachel Giora. 2003.
\newblock \emph{On our mind: Salience, context, and figurative language}.
\newblock Oxford University Press.

\bibitem[{Gopnik and Wellman(1992)}]{gopnik1992child}
Alison Gopnik and Henry~M Wellman. 1992.
\newblock Why the child's theory of mind really is a theory.
\newblock \emph{Mind \& Language}, 7(1--2):145--171.

\bibitem[{Gordon(2016)}]{gordon2016commonsense}
Andrew Gordon. 2016.
\newblock Commonsense interpretation of triangle behavior.
\newblock In \emph{Proceedings of the AAAI Conference on Artificial Intelligence}, volume~30.

\bibitem[{Grant et~al.(2017)Grant, Nematzadeh, and Griffiths}]{grant2017can}
Erin Grant, Aida Nematzadeh, and Thomas~L Griffiths. 2017.
\newblock How can memory-augmented neural networks pass a false-belief task?
\newblock In \emph{CogSci}.

\bibitem[{Gunning(2018)}]{gunning2018machine}
David Gunning. 2018.
\newblock Machine common sense concept paper.
\newblock \emph{arXiv preprint arXiv:1810.07528}.

\bibitem[{Hadwin et~al.(1997)Hadwin, Baron-Cohen, Howlin, and Hill}]{hadwin1997does}
Julie Hadwin, Simon Baron-Cohen, Patricia Howlin, and Katie Hill. 1997.
\newblock Does teaching theory of mind have an effect on the ability to develop conversation in children with autism?
\newblock \emph{Journal of autism and developmental disorders}, 27:519--537.

\bibitem[{Hagendorff(2023)}]{hagendorff2023machine}
Thilo Hagendorff. 2023.
\newblock Machine psychology: Investigating emergent capabilities and behavior in large language models using psychological methods.
\newblock \emph{arXiv preprint arXiv:2303.13988}.

\bibitem[{Happ{\'e}(1994)}]{happe1994advanced}
Francesca~GE Happ{\'e}. 1994.
\newblock An advanced test of theory of mind: Understanding of story characters' thoughts and feelings by able autistic, mentally handicapped, and normal children and adults.
\newblock \emph{Journal of autism and Developmental disorders}, 24(2):129--154.

\bibitem[{Harris et~al.(1986)Harris, Donnelly, Guz, and Pitt-Watson}]{harris1986children}
Paul~L Harris, Kara Donnelly, Gabrielle~R Guz, and Rosemary Pitt-Watson. 1986.
\newblock Children's understanding of the distinction between real and apparent emotion.
\newblock \emph{Child development}, pages 895--909.

\bibitem[{Hase et~al.(2023)Hase, Diab, Celikyilmaz, Li, Kozareva, Stoyanov, Bansal, and Iyer}]{hase2023methods}
Peter Hase, Mona Diab, Asli Celikyilmaz, Xian Li, Zornitsa Kozareva, Veselin Stoyanov, Mohit Bansal, and Srinivasan Iyer. 2023.
\newblock Methods for measuring, updating, and visualizing factual beliefs in language models.
\newblock In \emph{Proceedings of the 17th Conference of the European Chapter of the Association for Computational Linguistics}, pages 2706--2723.

\bibitem[{He et~al.(2023)He, Wu, Jia, Mihalcea, Chen, and Deng}]{he2023hi}
Yinghui He, Yufan Wu, Yilin Jia, Rada Mihalcea, Yulong Chen, and Naihao Deng. 2023.
\newblock Hi-tom: A benchmark for evaluating higher-order theory of mind reasoning in large language models.
\newblock In \emph{Findings of the Association for Computational Linguistics: EMNLP 2023}.

\bibitem[{Ho et~al.(2022)Ho, Saxe, and Cushman}]{ho2022planning}
Mark~K Ho, Rebecca Saxe, and Fiery Cushman. 2022.
\newblock Planning with theory of mind.
\newblock \emph{Trends in Cognitive Sciences}.

\bibitem[{Holterman and van Deemter(2023)}]{holterman2023does}
Bart Holterman and Kees van Deemter. 2023.
\newblock Does chatgpt have theory of mind?
\newblock \emph{arXiv preprint arXiv:2305.14020}.

\bibitem[{Hu et~al.(2022)Hu, Floyd, Jouravlev, Fedorenko, and Gibson}]{hu2022fine}
Jennifer Hu, Sammy Floyd, Olessia Jouravlev, Evelina Fedorenko, and Edward Gibson. 2022.
\newblock A fine-grained comparison of pragmatic language understanding in humans and language models.
\newblock \emph{arXiv preprint arXiv:2212.06801}.

\bibitem[{Jain et~al.(2019)Jain, Weihs, Kolve, Rastegari, Lazebnik, Farhadi, Schwing, and Kembhavi}]{jain2019two}
Unnat Jain, Luca Weihs, Eric Kolve, Mohammad Rastegari, Svetlana Lazebnik, Ali Farhadi, Alexander~G Schwing, and Aniruddha Kembhavi. 2019.
\newblock Two body problem: Collaborative visual task completion.
\newblock In \emph{Proceedings of the IEEE/CVF Conference on Computer Vision and Pattern Recognition}, pages 6689--6699.

\bibitem[{Jia et~al.(2022)Jia, He, Zhang, Uprety, Song, and Lioma}]{jia2022beyond}
Ao~Jia, Yu~He, Yazhou Zhang, Sagar Uprety, Dawei Song, and Christina Lioma. 2022.
\newblock Beyond emotion: A multi-modal dataset for human desire understanding.
\newblock In \emph{Proceedings of the 2022 Conference of the North American Chapter of the Association for Computational Linguistics: Human Language Technologies}, pages 1512--1522.

\bibitem[{Kavanagh et~al.(2005)Kavanagh, Andrade, and May}]{kavanagh2005imaginary}
David~J Kavanagh, Jackie Andrade, and Jon May. 2005.
\newblock Imaginary relish and exquisite torture: the elaborated intrusion theory of desire.
\newblock \emph{Psychological review}, 112(2):446.

\bibitem[{Kennington(2022)}]{kennington2022understanding}
Casey Kennington. 2022.
\newblock Understanding intention for machine theory of mind: a position paper.
\newblock In \emph{2022 31st IEEE International Conference on Robot and Human Interactive Communication (RO-MAN)}, pages 450--453. IEEE.

\bibitem[{Knafo et~al.(2009)Knafo, Zahn-Waxler, Davidov, Van~Hulle, Robinson, and Rhee}]{knafo2009empathy}
Ariel Knafo, Carolyn Zahn-Waxler, Maayan Davidov, Carol Van~Hulle, JoAnn~L Robinson, and Soo~Hyun Rhee. 2009.
\newblock Empathy in early childhood: Genetic, environmental, and affective contributions.
\newblock \emph{Annals of the New York Academy of Sciences}, 1167(1):103--114.

\bibitem[{Kosinski(2023)}]{kosinski2023theory}
Michal Kosinski. 2023.
\newblock Theory of mind may have spontaneously emerged in large language models.
\newblock \emph{arXiv preprint arXiv:2302.02083}.

\bibitem[{Kr{\"a}mer et~al.(2012)Kr{\"a}mer, P{\"u}tten, and Eimler}]{kramer2012human}
Nicole~C Kr{\"a}mer, Astrid von~der P{\"u}tten, and Sabrina Eimler. 2012.
\newblock Human-agent and human-robot interaction theory: Similarities to and differences from human-human interaction.
\newblock In \emph{Human-computer interaction: The agency perspective}, pages 215--240. Springer.

\bibitem[{Le et~al.(2019)Le, Boureau, and Nickel}]{le2019revisiting}
Matthew Le, Y-Lan Boureau, and Maximilian Nickel. 2019.
\newblock Revisiting the evaluation of theory of mind through question answering.
\newblock In \emph{Proceedings of the 2019 Conference on Empirical Methods in Natural Language Processing and the 9th International Joint Conference on Natural Language Processing (EMNLP-IJCNLP)}, pages 5872--5877, Hong Kong, China. Association for Computational Linguistics.

\bibitem[{Leer et~al.(2023)Leer, Trost, and Voruganti}]{leer2023violation}
Courtland Leer, Vincent Trost, and Vineeth Voruganti. 2023.
\newblock Violation of expectation via metacognitive prompting reduces theory of mind prediction error in large language models.
\newblock \emph{arXiv preprint arXiv:2310.06983}.

\bibitem[{Li et~al.(2021)Li, Nye, and Andreas}]{li2021implicit}
Belinda~Z Li, Maxwell Nye, and Jacob Andreas. 2021.
\newblock Implicit representations of meaning in neural language models.
\newblock In \emph{Proceedings of the 59th Annual Meeting of the Association for Computational Linguistics and the 11th International Joint Conference on Natural Language Processing (Volume 1: Long Papers)}, pages 1813--1827.

\bibitem[{Li et~al.(2023)Li, Hammoud, Itani, Khizbullin, and Ghanem}]{li2023camel}
Guohao Li, Hasan Abed Al~Kader Hammoud, Hani Itani, Dmitrii Khizbullin, and Bernard Ghanem. 2023.
\newblock Camel: Communicative agents for" mind" exploration of large scale language model society.
\newblock \emph{arXiv preprint arXiv:2303.17760}.

\bibitem[{Li et~al.(2022)Li, Puig, Paxton, Du, Wang, Fan, Chen, Huang, Aky{\"u}rek, Anandkumar et~al.}]{li2022pre}
Shuang Li, Xavier Puig, Chris Paxton, Yilun Du, Clinton Wang, Linxi Fan, Tao Chen, De-An Huang, Ekin Aky{\"u}rek, Anima Anandkumar, et~al. 2022.
\newblock Pre-trained language models for interactive decision-making.
\newblock \emph{Advances in Neural Information Processing Systems}, 35:31199--31212.

\bibitem[{Liu et~al.(2023)Liu, Zhu, Liu, Bisk, and Neubig}]{liu2023computational}
Andy Liu, Hao Zhu, Emmy Liu, Yonatan Bisk, and Graham Neubig. 2023.
\newblock \href {https://openreview.net/forum?id=C2ulri4duIs} {Computational language acquisition with theory of mind}.
\newblock In \emph{The Eleventh International Conference on Learning Representations}.

\bibitem[{Magar and Schwartz(2022)}]{magar2022data}
Inbal Magar and Roy Schwartz. 2022.
\newblock Data contamination: From memorization to exploitation.
\newblock In \emph{Proceedings of the 60th Annual Meeting of the Association for Computational Linguistics (Volume 2: Short Papers)}, pages 157--165.

\bibitem[{Malle and Knobe(2001)}]{malle2001distinction}
Bertram~F Malle and Joshua Knobe. 2001.
\newblock The distinction between desire and intention: A folk-conceptual analysis.
\newblock \emph{Intentions and intentionality: Foundations of social cognition}, 45:67.

\bibitem[{Marcus and Davis(2023)}]{marcus2023how}
Gary Marcus and Ernest Davis. 2023.
\newblock \href {https://garymarcus.substack.com/p/ how-not-to-test-gpt-3} {How not to test gpt-3}.

\bibitem[{Masangkay et~al.(1974)Masangkay, McCluskey, McIntyre, Sims-Knight, Vaughn, and Flavell}]{masangkay1974early}
Zenaida~S Masangkay, Kathleen~A McCluskey, Curtis~W McIntyre, Judith Sims-Knight, Brian~E Vaughn, and John~H Flavell. 1974.
\newblock The early development of inferences about the visual percepts of others.
\newblock \emph{Child development}, pages 357--366.

\bibitem[{Meltzoff(1995)}]{meltzoff1995understanding}
Andrew~N Meltzoff. 1995.
\newblock Understanding the intentions of others: re-enactment of intended acts by 18-month-old children.
\newblock \emph{Developmental psychology}, 31(5):838.

\bibitem[{Mitchell and Krakauer(2023)}]{mitchell2023debate}
Melanie Mitchell and David~C Krakauer. 2023.
\newblock The debate over understanding in ai’s large language models.
\newblock \emph{Proceedings of the National Academy of Sciences}, 120(13):e2215907120.

\bibitem[{Moghaddam and Honey(2023)}]{moghaddam2023boosting}
Shima~Rahimi Moghaddam and Christopher~J Honey. 2023.
\newblock Boosting theory-of-mind performance in large language models via prompting.
\newblock \emph{arXiv preprint arXiv:2304.11490}.

\bibitem[{Moll et~al.(2006)Moll, Koring, Carpenter, and Tomasello}]{moll2006infants}
Henrike Moll, Cornelia Koring, Malinda Carpenter, and Michael Tomasello. 2006.
\newblock Infants determine others' focus of attention by pragmatics and exclusion.
\newblock \emph{Journal of Cognition and Development}, 7(3):411--430.

\bibitem[{Nematzadeh et~al.(2018)Nematzadeh, Burns, Grant, Gopnik, and Griffiths}]{nematzadeh2018evaluating}
Aida Nematzadeh, Kaylee Burns, Erin Grant, Alison Gopnik, and Tom Griffiths. 2018.
\newblock Evaluating theory of mind in question answering.
\newblock In \emph{Proceedings of the 2018 Conference on Empirical Methods in Natural Language Processing}, pages 2392--2400, Brussels, Belgium. Association for Computational Linguistics.

\bibitem[{OpenAI(2022)}]{chatgpt2022openai}
OpenAI. 2022.
\newblock \href {https://openai.com/blog/chatgpt/} {Chatgpt: Optimizing language models for dialogue}.

\bibitem[{OpenAI(2023)}]{openai2023gpt4}
OpenAI. 2023.
\newblock Gpt-4 technical report.
\newblock \emph{arXiv preprint arXiv:2303.08774}.

\bibitem[{Osterhaus and Bosacki(2022)}]{osterhaus2022looking}
Christopher Osterhaus and Sandra~L Bosacki. 2022.
\newblock Looking for the lighthouse: A systematic review of advanced theory-of-mind tests beyond preschool.
\newblock \emph{Developmental Review}, 64:101021.

\bibitem[{Pereira et~al.(2016)Pereira, Prada, and Santos}]{pereira2016integrating}
Gon{\c{c}}alo Pereira, Rui Prada, and Pedro~A Santos. 2016.
\newblock Integrating social power into the decision-making of cognitive agents.
\newblock \emph{Artificial Intelligence}, 241:1--44.

\bibitem[{Perner et~al.(1987)Perner, Leekam, and Wimmer}]{perner1987three}
Josef Perner, Susan~R Leekam, and Heinz Wimmer. 1987.
\newblock Three-year-olds' difficulty with false belief: The case for a conceptual deficit.
\newblock \emph{British journal of developmental psychology}, 5(2):125--137.

\bibitem[{Perner and Wimmer(1985)}]{perner1985john}
Josef Perner and Heinz Wimmer. 1985.
\newblock “john thinks that mary thinks that…” attribution of second-order beliefs by 5-to 10-year-old children.
\newblock \emph{Journal of experimental child psychology}, 39(3):437--471.

\bibitem[{Phillips et~al.(2002)Phillips, Wellman, and Spelke}]{phillips2002infants}
Ann~T Phillips, Henry~M Wellman, and Elizabeth~S Spelke. 2002.
\newblock Infants' ability to connect gaze and emotional expression to intentional action.
\newblock \emph{Cognition}, 85(1):53--78.

\bibitem[{Pons and Harris(2000)}]{pons2000test}
F.~Pons and P.~Harris. 2000.
\newblock \emph{Test of Emotion Comprehension: TEC}.
\newblock University of Oxford.

\bibitem[{Premack and Woodruff(1978)}]{premack1978chimpanzee}
David Premack and Guy Woodruff. 1978.
\newblock Does the chimpanzee have a theory of mind?
\newblock \emph{Behavioral and Brain Sciences}, 1(4):515–526.

\bibitem[{Qiu et~al.(2022)Qiu, Zhao, Liang, Lu, Shi, Yu, and Zhu}]{qiu2022towards}
Liang Qiu, Yizhou Zhao, Yuan Liang, Pan Lu, Weiyan Shi, Zhou Yu, and Song-chun Zhu. 2022.
\newblock Towards socially intelligent agents with mental state transition and human value.
\newblock In \emph{Proceedings of the 23rd Annual Meeting of the Special Interest Group on Discourse and Dialogue}, pages 146--158.

\bibitem[{Rabinowitz et~al.(2018)Rabinowitz, Perbet, Song, Zhang, Eslami, and Botvinick}]{rabinowitz2018machine}
Neil Rabinowitz, Frank Perbet, Francis Song, Chiyuan Zhang, SM~Ali Eslami, and Matthew Botvinick. 2018.
\newblock Machine theory of mind.
\newblock In \emph{International conference on machine learning}, pages 4218--4227. PMLR.

\bibitem[{Raji et~al.(2021)Raji, Denton, Bender, Hanna, and Paullada}]{raji2ai}
Inioluwa~Deborah Raji, Emily Denton, Emily~M Bender, Alex Hanna, and Amandalynne Paullada. 2021.
\newblock Ai and the everything in the whole wide world benchmark.
\newblock In \emph{Thirty-fifth Conference on Neural Information Processing Systems Datasets and Benchmarks Track (Round 2)}.

\bibitem[{Repacholi and Gopnik(1997)}]{repacholi1997early}
Betty~M Repacholi and Alison Gopnik. 1997.
\newblock Early reasoning about desires: evidence from 14-and 18-month-olds.
\newblock \emph{Developmental psychology}, 33(1):12.

\bibitem[{Ruffman and Olson(1989)}]{ruffman1989children}
Ted~K Ruffman and David~R Olson. 1989.
\newblock Children's ascriptions of knowledge to others.
\newblock \emph{Developmental Psychology}, 25(4):601.

\bibitem[{Ruis et~al.(2022)Ruis, Khan, Biderman, Hooker, Rockt{\"a}schel, and Grefenstette}]{ruis2022large}
Laura Ruis, Akbir Khan, Stella Biderman, Sara Hooker, Tim Rockt{\"a}schel, and Edward Grefenstette. 2022.
\newblock Large language models are not zero-shot communicators.
\newblock \emph{arXiv preprint arXiv:2210.14986}.

\bibitem[{Rusch et~al.(2020)Rusch, Steixner-Kumar, Doshi, Spezio, and Gl{\"a}scher}]{rusch2020theory}
Tessa Rusch, Saurabh Steixner-Kumar, Prashant Doshi, Michael Spezio, and Jan Gl{\"a}scher. 2020.
\newblock Theory of mind and decision science: towards a typology of tasks and computational models.
\newblock \emph{Neuropsychologia}, 146:107488.

\bibitem[{Sap et~al.(2022)Sap, Le~Bras, Fried, and Choi}]{sap2022neural}
Maarten Sap, Ronan Le~Bras, Daniel Fried, and Yejin Choi. 2022.
\newblock Neural theory-of-mind? on the limits of social intelligence in large {LM}s.
\newblock In \emph{Proceedings of the 2022 Conference on Empirical Methods in Natural Language Processing}, pages 3762--3780, Abu Dhabi, United Arab Emirates. Association for Computational Linguistics.

\bibitem[{Sap et~al.(2019)Sap, Rashkin, Chen, Le~Bras, and Choi}]{sap2019social}
Maarten Sap, Hannah Rashkin, Derek Chen, Ronan Le~Bras, and Yejin Choi. 2019.
\newblock Social {IQ}a: Commonsense reasoning about social interactions.
\newblock In \emph{Proceedings of the 2019 Conference on Empirical Methods in Natural Language Processing and the 9th International Joint Conference on Natural Language Processing (EMNLP-IJCNLP)}, pages 4463--4473, Hong Kong, China. Association for Computational Linguistics.

\bibitem[{Schank and Abelson(2013)}]{schank2013scripts}
Roger~C Schank and Robert~P Abelson. 2013.
\newblock \emph{Scripts, plans, goals, and understanding: An inquiry into human knowledge structures}.
\newblock Psychology press.

\bibitem[{Schulman et~al.(2017)Schulman, Wolski, Dhariwal, Radford, and Klimov}]{schulman2017proximal}
John Schulman, Filip Wolski, Prafulla Dhariwal, Alec Radford, and Oleg Klimov. 2017.
\newblock Proximal policy optimization algorithms.
\newblock \emph{arXiv preprint arXiv:1707.06347}.

\bibitem[{Sclar et~al.(2023)Sclar, Kumar, West, Suhr, Choi, and Tsvetkov}]{sclar2023minding}
Melanie Sclar, Sachin Kumar, Peter West, Alane Suhr, Yejin Choi, and Yulia Tsvetkov. 2023.
\newblock \href {http://arxiv.org/abs/2306.00924} {Minding language models' (lack of) theory of mind: A plug-and-play multi-character belief tracker}.

\bibitem[{Sclar et~al.(2022)Sclar, Neubig, and Bisk}]{sclar2022symmetric}
Melanie Sclar, Graham Neubig, and Yonatan Bisk. 2022.
\newblock Symmetric machine theory of mind.
\newblock In \emph{Proceedings of the 39th International Conference on Machine Learning}, volume 162, pages 19450--19466.

\bibitem[{Shapira et~al.(2023{\natexlab{a}})Shapira, Levy, Alavi, Zhou, Choi, Goldberg, Sap, and Shwartz}]{shapira2023clever}
Natalie Shapira, Mosh Levy, Seyed~Hossein Alavi, Xuhui Zhou, Yejin Choi, Yoav Goldberg, Maarten Sap, and Vered Shwartz. 2023{\natexlab{a}}.
\newblock \href {http://arxiv.org/abs/2305.14763} {Clever hans or neural theory of mind? stress testing social reasoning in large language models}.

\bibitem[{Shapira et~al.(2023{\natexlab{b}})Shapira, Zwirn, and Goldberg}]{shapira2023how}
Natalie Shapira, Guy Zwirn, and Yoav Goldberg. 2023{\natexlab{b}}.
\newblock How well do large language models perform on faux pas tests.
\newblock In \emph{Findings of the Association for Computational Linguistics: ACL 2023}.

\bibitem[{Shu et~al.(2021)Shu, Bhandwaldar, Gan, Smith, Liu, Gutfreund, Spelke, Tenenbaum, and Ullman}]{shu2021agent}
Tianmin Shu, Abhishek Bhandwaldar, Chuang Gan, Kevin Smith, Shari Liu, Dan Gutfreund, Elizabeth Spelke, Joshua Tenenbaum, and Tomer Ullman. 2021.
\newblock Agent: A benchmark for core psychological reasoning.
\newblock In \emph{International Conference on Machine Learning}, pages 9614--9625. PMLR.

\bibitem[{Sileo and Lernould(2023)}]{sileo2023mindgames}
Damien Sileo and Antoine Lernould. 2023.
\newblock Mindgames: Targeting theory of mind in large language models with dynamic epistemic modal logic.
\newblock \emph{arXiv preprint arXiv:2305.03353}.

\bibitem[{Smiley(2001)}]{smiley2001intention}
Patricia~A Smiley. 2001.
\newblock Intention understanding and partner-sensitive behaviors in young children’s peer interactions.
\newblock \emph{Social Development}, 10(3):330--354.

\bibitem[{Stack et~al.(2022)Stack, Farhana, Shen, Zhao, and Maliakal}]{stack2022framework}
Caoimhe~Harrington Stack, Effat Farhana, Xinyu Shen, Simeng Zhao, and Angela Maliakal. 2022.
\newblock Framework for a multi-dimensional test of theory of mind for humans and ai systems.
\newblock In \emph{The Tenth Annual Conference on Advances in Cognitive Systems}.

\bibitem[{Storks et~al.(2021)Storks, Gao, Zhang, and Chai}]{storks2021tiered}
Shane Storks, Qiaozi Gao, Yichi Zhang, and Joyce Chai. 2021.
\newblock Tiered reasoning for intuitive physics: Toward verifiable commonsense language understanding.
\newblock In \emph{Findings of the Association for Computational Linguistics: EMNLP 2021}, pages 4902--4918.

\bibitem[{Sullivan et~al.(1995)Sullivan, Winner, and Hopfield}]{sullivan1995children}
Kate Sullivan, Ellen Winner, and Natalie Hopfield. 1995.
\newblock How children tell a lie from a joke: The role of second-order mental state attributions.
\newblock \emph{British journal of developmental psychology}, 13(2):191--204.

\bibitem[{Swettenham(1996)}]{swettenham1996can}
J~Swettenham. 1996.
\newblock Can children be taught to understand false belief using computers? child psychology \& psychiatry \& allied disciplines, 37 (2), 157--165.

\bibitem[{Takmaz et~al.(2023)Takmaz, Brandizzi, Giulianelli, Pezzelle, and Fern{\'a}ndez}]{takmaz2023speaking}
Ece Takmaz, Nicolo' Brandizzi, Mario Giulianelli, Sandro Pezzelle, and Raquel Fern{\'a}ndez. 2023.
\newblock Speaking the language of your listener: Audience-aware adaptation via plug-and-play theory of mind.
\newblock In \emph{Findings of the Association for Computational Linguistics: ACL 2023}.

\bibitem[{Tomasello(2005)}]{tomasello2005constructing}
Michael Tomasello. 2005.
\newblock \emph{Constructing a language: A usage-based theory of language acquisition}.
\newblock Harvard university press.

\bibitem[{Tracey et~al.(2022)Tracey, Rambow, Cardie, Dalton, Dang, Diab, Dorr, Guthrie, Markowska, Muresan et~al.}]{tracey2022best}
Jennifer Tracey, Owen Rambow, Claire Cardie, Adam Dalton, Hoa~Trang Dang, Mona Diab, Bonnie Dorr, Louise Guthrie, Magdalena Markowska, Smaranda Muresan, et~al. 2022.
\newblock Best: The belief and sentiment corpus.
\newblock In \emph{Proceedings of the Thirteenth Language Resources and Evaluation Conference}, pages 2460--2467.

\bibitem[{Trott et~al.(2022)Trott, Jones, Chang, Michaelov, and Bergen}]{trott2022large}
Sean Trott, Cameron Jones, Tyler Chang, James Michaelov, and Benjamin Bergen. 2022.
\newblock Do large language models know what humans know?
\newblock \emph{arXiv preprint arXiv:2209.01515}.

\bibitem[{Ullman(2023)}]{ullman2023large}
Tomer Ullman. 2023.
\newblock Large language models fail on trivial alterations to theory-of-mind tasks.
\newblock \emph{arXiv preprint arXiv:2302.08399}.

\bibitem[{Wang et~al.(2021)Wang, Saha, Gregori, Joyner, and Goel}]{wang2021towards}
Qiaosi Wang, Koustuv Saha, Eric Gregori, David Joyner, and Ashok Goel. 2021.
\newblock Towards mutual theory of mind in human-ai interaction: How language reflects what students perceive about a virtual teaching assistant.
\newblock In \emph{Proceedings of the 2021 CHI Conference on Human Factors in Computing Systems}, pages 1--14.

\bibitem[{Wang et~al.(2023)Wang, Zhang, Yang, Shi, Zhou, Hao, Xiong, Li, Sim, Chen et~al.}]{wang2023interactive}
Zekun Wang, Ge~Zhang, Kexin Yang, Ning Shi, Wangchunshu Zhou, Shaochun Hao, Guangzheng Xiong, Yizhi Li, Mong~Yuan Sim, Xiuying Chen, et~al. 2023.
\newblock Interactive natural language processing.
\newblock \emph{arXiv preprint arXiv:2305.13246}.

\bibitem[{Wei et~al.(2022)Wei, Wang, Schuurmans, Bosma, Xia, Chi, Le, Zhou et~al.}]{wei2022chain}
Jason Wei, Xuezhi Wang, Dale Schuurmans, Maarten Bosma, Fei Xia, Ed~H Chi, Quoc~V Le, Denny Zhou, et~al. 2022.
\newblock Chain-of-thought prompting elicits reasoning in large language models.
\newblock In \emph{Advances in Neural Information Processing Systems}.

\bibitem[{Wellman and Woolley(1990)}]{wellman1990simple}
Henry~M Wellman and Jacqueline~D Woolley. 1990.
\newblock From simple desires to ordinary beliefs: The early development of everyday psychology.
\newblock \emph{Cognition}, 35(3):245--275.

\bibitem[{Wu et~al.(2023)Wu, Chen, Deng, Sabour, and Huang}]{wu2023coke}
Jincenzi Wu, Zhuang Chen, Jiawen Deng, Sahand Sabour, and Minlie Huang. 2023.
\newblock Coke: A cognitive knowledge graph for machine theory of mind.
\newblock \emph{arXiv preprint arXiv:2305.05390}.

\bibitem[{Yoshida et~al.(2008)Yoshida, Dolan, and Friston}]{yoshida2008game}
Wako Yoshida, Ray~J Dolan, and Karl~J Friston. 2008.
\newblock Game theory of mind.
\newblock \emph{PLoS computational biology}, 4(12):e1000254.

\bibitem[{Yu et~al.(2022)Yu, Sang, Pu, Wei, Wang, Li, Yu, and Zhou}]{yu2022few}
Mo~Yu, Yisi Sang, Kangsheng Pu, Zekai Wei, Han Wang, Jing Li, Yue Yu, and Jie Zhou. 2022.
\newblock Few-shot character understanding in movies as an assessment to meta-learning of theory-of-mind.
\newblock \emph{arXiv preprint arXiv:2211.04684}.

\bibitem[{Zaki et~al.(2009)Zaki, Bolger, and Ochsner}]{zaki2009unpacking}
Jamil Zaki, Niall Bolger, and Kevin Ochsner. 2009.
\newblock Unpacking the informational bases of empathic accuracy.
\newblock \emph{Emotion}, 9(4):478.

\bibitem[{Zhang and Chai(2010)}]{zhang2010towards}
Chen Zhang and Joyce~Y Chai. 2010.
\newblock Towards conversation entailment: An empirical investigation.
\newblock In \emph{Proceedings of the 2010 Conference on Empirical Methods in Natural Language Processing}, pages 756--766.

\bibitem[{Zhang et~al.(2021)Zhang, Zhang, Zhan, Chen, Shi, Wu, and Lam}]{zhang2021effectiveness}
Haode Zhang, Yuwei Zhang, Li~Ming Zhan, Jiaxin Chen, Guangyuan Shi, Xiao~Ming Wu, and Albert~YS Lam. 2021.
\newblock Effectiveness of pre-training for few-shot intent classification.
\newblock In \emph{2021 Findings of the Association for Computational Linguistics, Findings of ACL: EMNLP 2021}, pages 1114--1120. Association for Computational Linguistics (ACL).

\bibitem[{Zhou et~al.(2023{\natexlab{a}})Zhou, Madaan, Potharaju, Gupta, McKee, Holtzman, Pujara, Ren, Mishra, Nematzadeh et~al.}]{zhou2023far}
Pei Zhou, Aman Madaan, Srividya~Pranavi Potharaju, Aditya Gupta, Kevin~R McKee, Ari Holtzman, Jay Pujara, Xiang Ren, Swaroop Mishra, Aida Nematzadeh, et~al. 2023{\natexlab{a}}.
\newblock How far are large language models from agents with theory-of-mind?
\newblock \emph{arXiv preprint arXiv:2310.03051}.

\bibitem[{Zhou et~al.(2023{\natexlab{b}})Zhou, Zhu, Hu, Pujara, Ren, Callison-Burch, Choi, and Ammanabrolu}]{zhou2023i}
Pei Zhou, Andrew Zhu, Jennifer Hu, Jay Pujara, Xiang Ren, Chris Callison-Burch, Yejin Choi, and Prithviraj Ammanabrolu. 2023{\natexlab{b}}.
\newblock I cast detect thoughts: Learning to converse and guide with intents and theory-of-mind in dungeons and dragons.
\newblock In \emph{Proceedings of the 61th Annual Meeting of the Association for Computational Linguistics}.

\bibitem[{Zhu et~al.(2021{\natexlab{a}})Zhu, Neubig, and Bisk}]{zhu2021few}
Hao Zhu, Graham Neubig, and Yonatan Bisk. 2021{\natexlab{a}}.
\newblock Few-shot language coordination by modeling theory of mind.
\newblock In \emph{International Conference on Machine Learning}, pages 12901--12911. PMLR.

\bibitem[{Zhu et~al.(2021{\natexlab{b}})Zhu, Neubig, and Bisk}]{zhu2021fewshot}
Hao Zhu, Graham Neubig, and Yonatan Bisk. 2021{\natexlab{b}}.
\newblock Few-shot language coordination by modeling theory of mind.
\newblock In \emph{Proceedings of the 38th International Conference on Machine Learning}, volume 139 of \emph{Proceedings of Machine Learning Research}, pages 12901--12911. PMLR.

\end{thebibliography}

\clearpage
\appendix

\section{Task Settings and Data Collection}
\label{app::setting}

In this section, we provide an in-depth explanation of the ten tasks outlined in section ~\ref{sec::minigrid}. Task 0 serves as a "reality check" to assess LLMs' grasp of the physical world, particularly relocations within the grid world. Tasks 1 through 9 each emphasize distinct facets of ToM. All these tasks utilize MiniGrid, a streamlined 2D grid-world environment.~\citep{minigrid}.

\subsubsection*{Task 0: Reality Check}

\begin{figure}[hp]
    \centering
    \includegraphics[width=.95\linewidth]{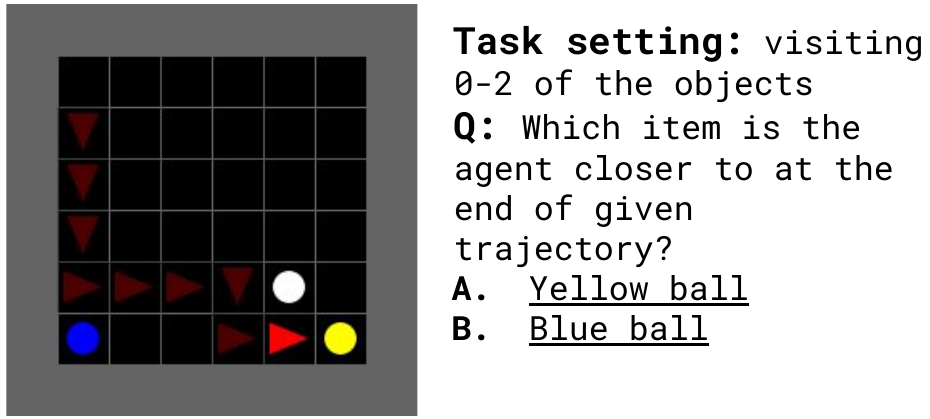}
    \vspace{-5pt}
    \caption{Illustration for \textit{Reality Check} task}
    \label{fig:task0}
\end{figure}

In this scenario, the agent is tasked with visiting 0-2 of the objects in the grid world. 
The agent has full observation in the world for efficient navigation. 
At least 2 objects (12 maximum) are placed in the environment. 
To test an LLM's ability to understand the physical actions taken by the agent, we ask it about the distance between the agent and various objects after it accomplishes a number of actions. 
The action planner is either a shortest path planner towards specified objects or a random action generator. 

After the agent has finished 10 random actions, or right after it has visited one object with an optimal action planner, the task-related question will be generated with the following format:

\vspace*{5pt}
\texttt{\textbf{After having taken these actions, which item is the agent closer to?}}
\vspace*{-5pt}
\begin{enumerate}[A., itemsep=-0.2em]
    \item \texttt{<object1.color> <object1.name>}
    \item \texttt{<object2.color> <object2.name>}
\end{enumerate}

Here \texttt{object1} and \texttt{object2} both exist in the environment, and one of them is guaranteed to be the target object if there is such a goal. 
There are always two options for this task. 

Along with the task-related question, the prompt includes a description of the grid world environment, the action space of the agent (only going forward, turn left / right in this task), a board-like depiction of the initial state, the list of actions taken by the agent, and the agent's location and face direction for each step. 
For more details on prompting, please refer to section~\ref{sec::prompting}.

The data for Task 0 are autonomously generated using seeds and a shortest-path planner.

\subsubsection*{Task 1: Short Term Intention}

\begin{figure}[hbtp]
    \centering
    \includegraphics[width=.95\linewidth]{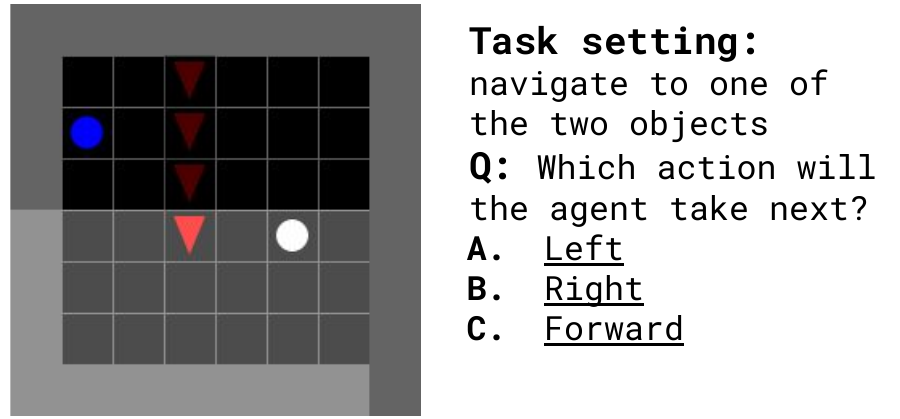}
    \vspace{-5pt}
    \caption{Illustration for \textit{Short Term Intention} task}
    \label{fig:task1}
\end{figure}

In this scenario, the agent is tasked with visiting either of the two objects in the grid world.
The LLM is not provided with the goal object.
Rather, it must determine the goal object by examining and understanding the agent's trajectory.
In this task, the agent has full observation in the world, and there are exactly two objects placed. 
The object types and colors are randomly generated. 
The size of the room can be randomly sampled from the range 6 by 6 to 12 by 12. 

To test an LLM's understanding of short-term intention, we ask it to predict the next action of the agent given its previous trajectory. 

The agent's trajectory halts at a random step, with the exception of the precise moment when the optimal paths to the two objects diverge. 
This "cutting point" is set as an exception and also serves as the mean for the normal distribution from which the stopping point is sampled.

By restricting the cutting point in this manner, we guarantee the trajectory included in the prompt to be optimal for reaching the potential goal objects.
This reduces the ambiguities of our experiments, and thus improves the significance of our results.

The task-related question has the following format:

\vspace*{5pt}
\texttt{\textbf{Which action will the agent take next?}}
\vspace*{-5pt}
\begin{enumerate}[A., itemsep=-0.2em]
    \item \texttt{left}
    \item \texttt{right}
    \item \texttt{forward}
\end{enumerate}

The LLM should be able to choose which action would be next were the agent to continue its optimal trajectory to the goal object.
The data for Task 1 are autonomously generated using seeds and a shortest-path planner.

\subsubsection*{Task 2: Long Term Intention}

\begin{figure}[hbtp]
    \centering
    \includegraphics[width=.9\linewidth]{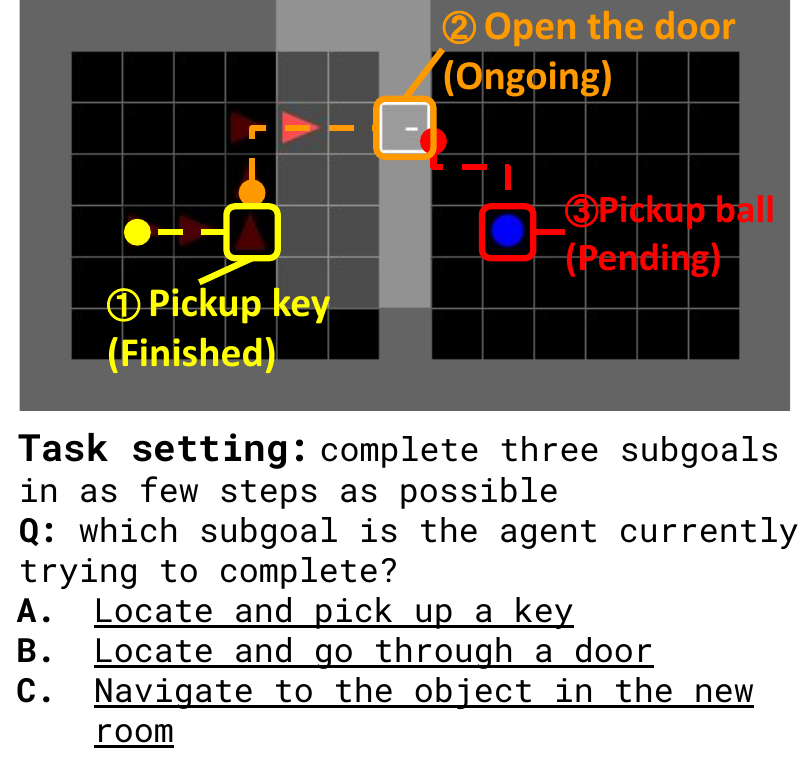}
    \vspace{-5pt}
    \caption{Illustration for \textit{Long term intention} task}
    \label{fig:task2}
\end{figure}

In this scenario, the agent needs to complete the following subgoals in as few steps as possible: 1) Locate and pick up a key; 2) Locate and go through a door; 3) Navigate to the object in the new room. 
There are two rooms in this setting, which are connected by a locked door. 
The key of the door is always in the room in which the agent is initially located.
The object is always placed in the other room.

We provide an LLM with a subset of the agent's trajectory and ask it which subgoal the agent is currently trying to complete. 

The task-related question has the following format:

\vspace*{5pt}
\texttt{\textbf{Based on the agent's trajectory thus far, which subgoal is the agent currently trying to complete?}}
\vspace*{-5pt}
\begin{enumerate}[A., itemsep=-0.2em]
    \item \texttt{Locate and pick up a key}
    \item \texttt{Locate and go through a door}
    \item \texttt{Navigate to the object in the new room}
\end{enumerate}

The data for Task 2 are autonomously generated using seeds and a shortest-path planner.

\subsubsection*{Task 3: Desire}

\begin{figure}[hbtp]
    \centering
    \includegraphics[width=.8\linewidth]{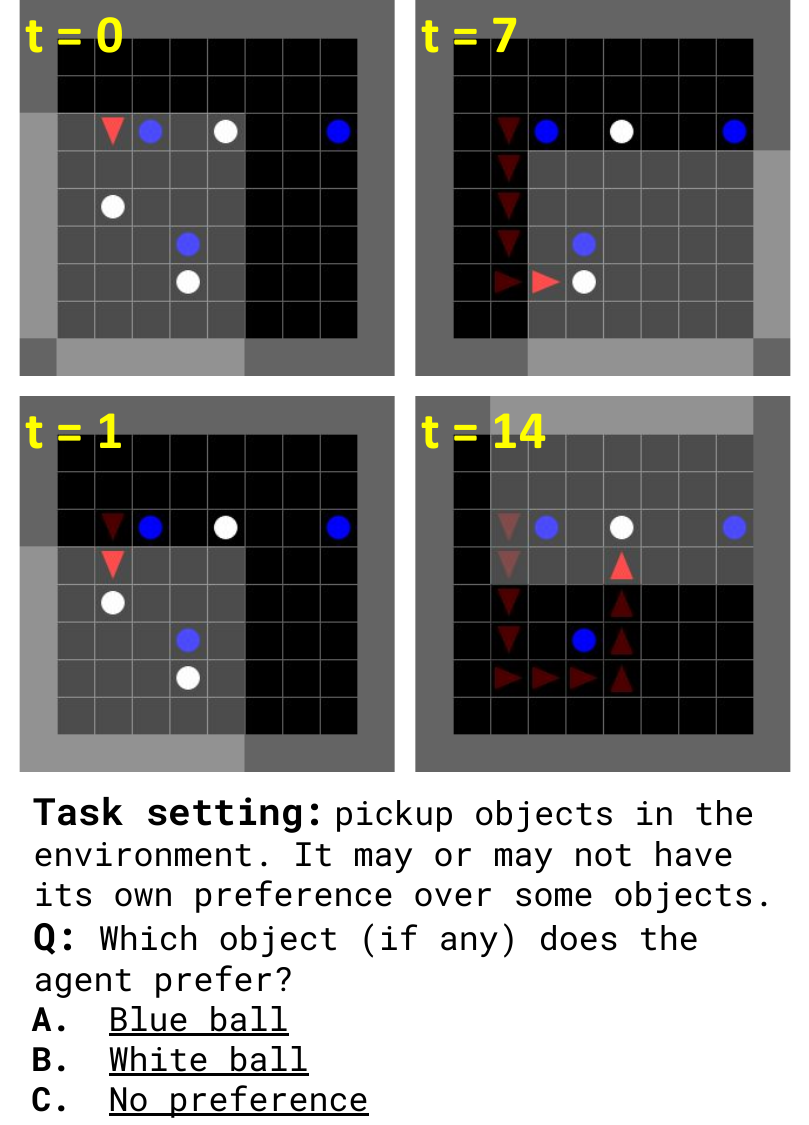}
    \vspace{-5pt}
    \caption{Illustration for \textit{Desire} task}
    \label{fig:task3}
\end{figure}

In this scenario, the agent is required to pick up three objects as soon as possible. 
There are 2 types of objects in the world (e.g., blue balls and white balls), 3 of each. 
The agent may or may not have a preference for one object type (we stratify the data such that in 50\% of the episodes the agent has a preferred object type and in the other 50\% the agent lacks a preference). 
We also deduct from the final reward given to the agent when it takes a large number of steps to finish picking up three objects. 

This task tests whether an LLM is able to determine the desire of the agent (for one object type or the other) by examining its trajectory.
We prepared the following question for this task:

\vspace*{5pt}
\texttt{\textbf{Which object does the agent prefer?}}
\vspace*{-5pt}
\begin{enumerate}[A., itemsep=-0.2em]
    \item \texttt{white ball}
    \item \texttt{blue ball}
    \item \texttt{no preference}
\end{enumerate}

The data for Task 3 are autonomously generated using seeds and Reinforcement Learning. 
We use the PPO algorithm \citep{schulman2017proximal} to train the model.
In the scenario wherein a preference is present, the preferred object type yields 10 times more reward than the non-preferred one.
In the scenario wherein the preference is absent, both object types yield identical rewards.

\subsubsection*{Task 4: Percept}

\begin{figure}[hbtp]
    \centering
    \hspace{-2pt}
    \includegraphics[width=1.02\linewidth]{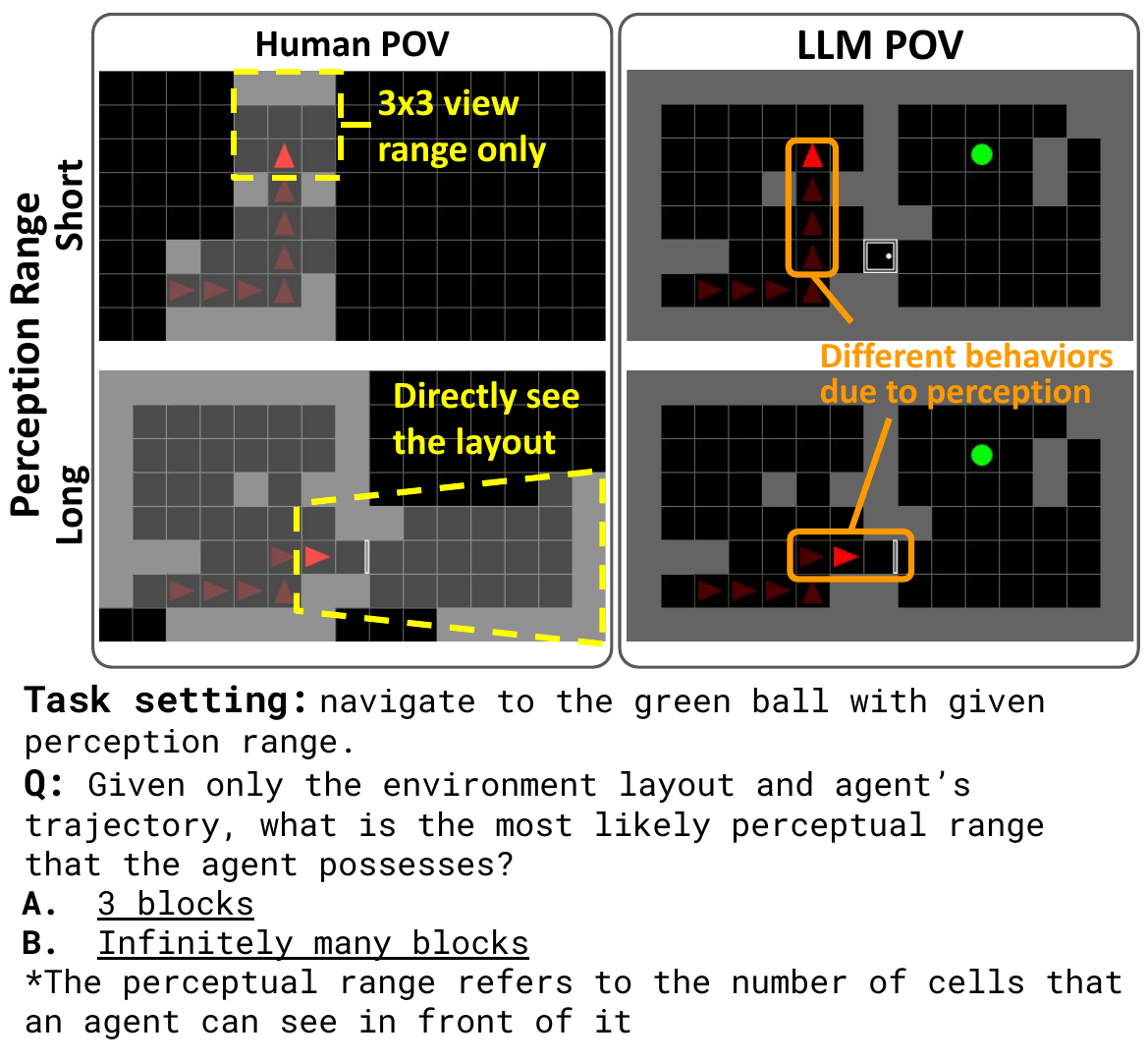}
    \vspace{-10pt}
    \caption{Illustration for \textit{Percept} task}
    \label{fig:task4}
\end{figure}

In this scenario, the agent is instructed to navigate in two rooms and reach the goal in the other room as fast as possible. 
In contrast with previous task settings, the agent has either a very limited visual range (3 x 3 grid in the front), or an ``infinitely'' large visual range (for practical purposes, the visual range is actually a 101 x 101 grid). 
Naturally, an agent with a smaller viewing range will make more mistakes (e.g., navigating to a dead end) while trying to reach the goal object.
Obstacles are randomly placed in each room to block the agent's view.

LLMs are expected to only look at the trajectory of the agent in an environment, and determine whether the agent has a limited view range or a nearly full view range. 
The question format is as follows:

\vspace*{5pt}
\texttt{\textbf{Based on the agent's actions, what is the most likely perceptual range that the agent possesses?
The perceptual range refers to the number of cells that an agent can see in front of it.}}
\vspace*{-5pt}
\begin{enumerate}[A., itemsep=-0.2em]
    \item \texttt{3 blocks}
    \item \texttt{infinitely many blocks}
\end{enumerate}

We manually collected 100 trials in total for both situations.

\subsubsection*{Task 5 \& 6: First and Second Order Belief}

In this scenario, there are two agents in the environment. 
Both of them are initially in the main room (on the left side of the whole grid world; see Figure ~\ref{fig:task-belief}). 
On the right side, there are three small rooms. Agents can freely go in and out of each room. 
They can see everything inside the current room and can see the other rooms through the door if it is open. 

This task reproduces the unexpected transfer (Sally-Anne) test ~\citep{baron1985does,perner1985john}, with both first-order and second-order belief checking. 
In the first-order belief task, one agent does not see the second agent transfer a ball from one room to another.
Presumably, therefore, the agent falsely believes the ball to be in its previous location.
The second-order belief task extends the first-order belief task by enabling the agent with the false belief to see the ball in its new location.
The other agent, however, does not witness the first agent rectifying its false belief, so it presumably holds a second-order false belief (about the belief state of the first agent).
By varying the observations that each agent makes (e.g., whether or not each agent sees the transfer of the ball from one room to the another) as well as the order of these events, this task setting allows us to check an LLM's first-order and second-order belief capabilities.

Rather than asking the LLM directly where to find the object, we provide two board-like belief states as options.
We then query the LLM about which board-like state the agent is more likely to believe.

Data for tasks 5 \& 6 are collected via rule-based planners with several scenarios.

\subsubsection*{Task 7: Non-Literal Communication}

\begin{figure}[hbtp]
    \centering
    \includegraphics[width=.9\linewidth]{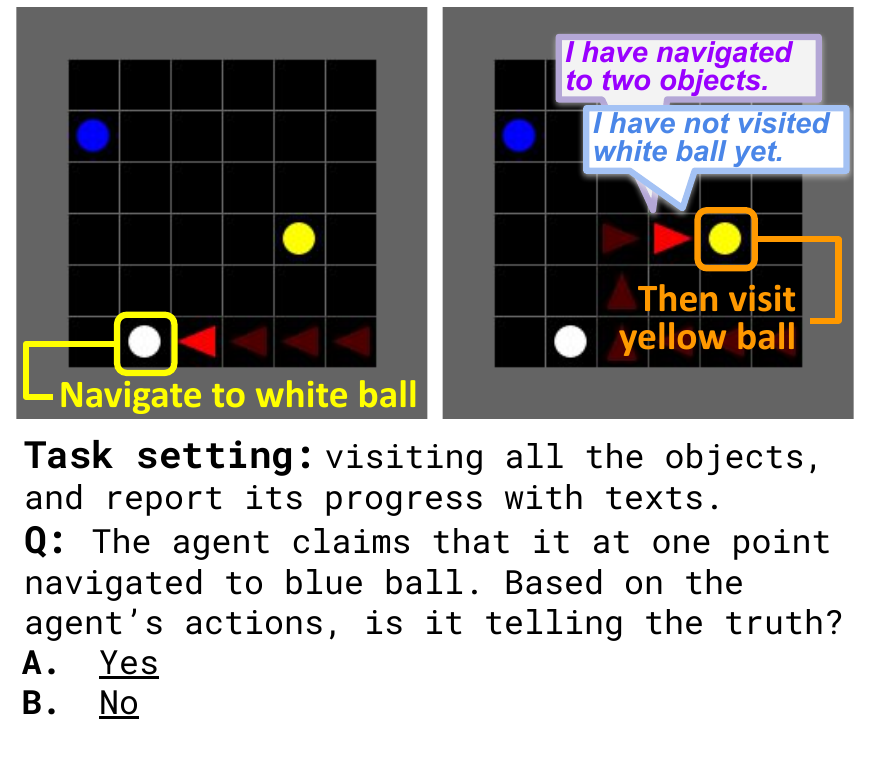}
    \caption{Illustration for \textit{Non-literal communication} task}
    \label{fig:task7}
\end{figure}

This task focuses specifically on one form of non-literal communication: lying. 
Within this task, the LLM is told explicitly that there is an agent tasked with navigating to all of the objects within the grid world. 
In each instance of the task, however, the agent only visits a subset of the objects in the environment. 
The LLM is subsequently told that the agent has claimed success in visiting a particular object. 
This object is randomly selected so that sometimes it is an object that has actually been visited and other times it is not.

The question format is as follows:

\vspace*{5pt}
\texttt{\textbf{Based on the agent's actions, is it telling the truth?}}
\vspace*{-5pt}
\begin{enumerate}[A., itemsep=-0.2em]
    \item \texttt{yes}
    \item \texttt{no}
\end{enumerate}

To successfully complete this task, the LLM must combine its knowledge about the physical occurrences taking place within the grid world with its knowledge about lying, a vital component in \citet{beaudoin2020systematic}'s category of non-literal communication.

\subsubsection*{Task 8: Knowledge}

\begin{figure}[hbtp]
    \centering
    \includegraphics[width=.95\linewidth]{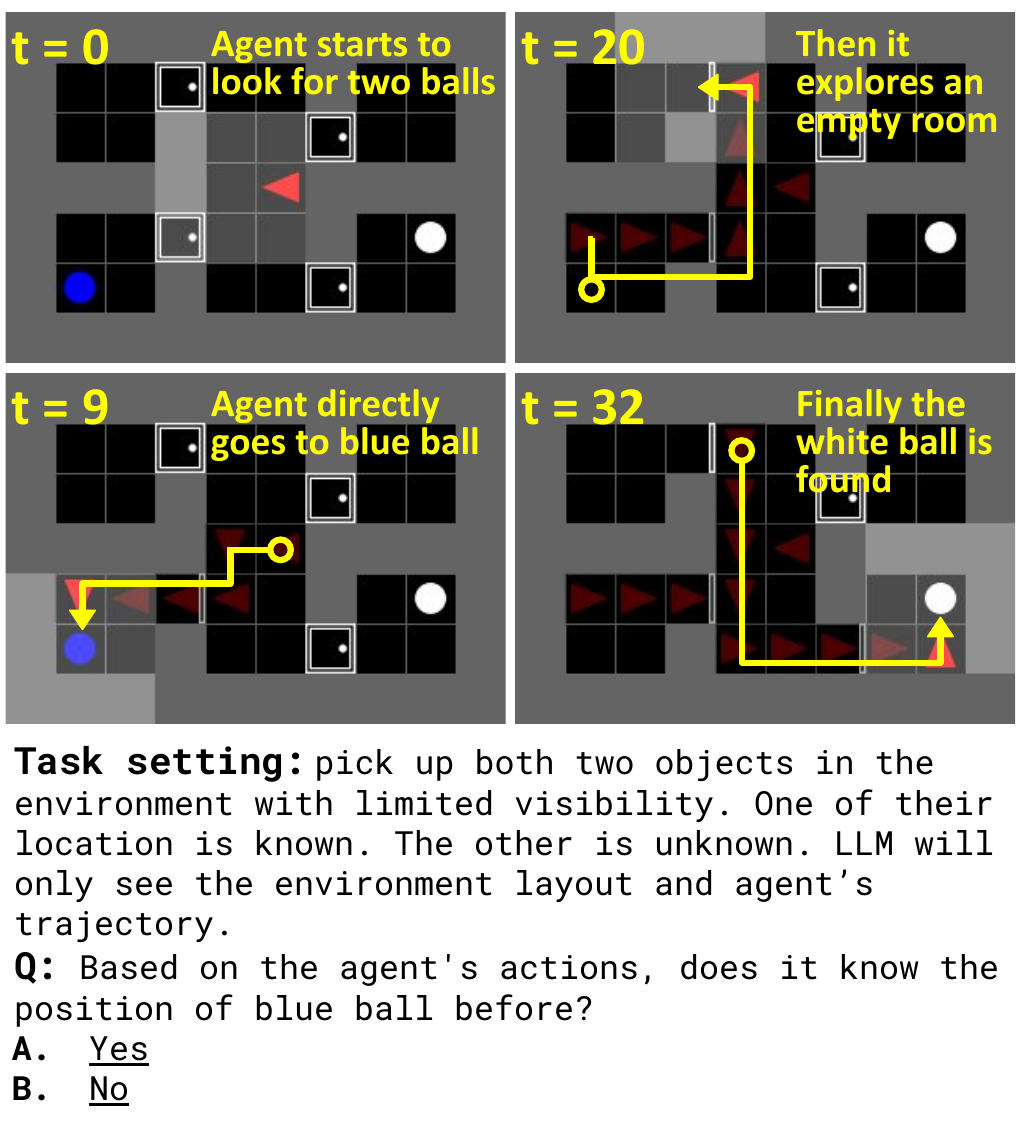}
    \caption{Illustration for \textit{Knowledge} task}
    \label{fig:task8}
\end{figure}

This task requires the agent to pick up two objects that exist in the environment. Both two objects are known to be placed separately in two of the four rooms. 

In this scenario, the agent is informed of the location of one object, while the location of the second object remains unknown. The agent is instructed to collect the objects in a specific sequence. Ideally, if the agent knows an object's location, it should proceed directly to the appropriate room. Otherwise, it will have to search for the object in the yet-to-be-explored rooms.

We include the agent's entire trajectory in the prompt.
We then ask the LLM to determine if the agent knows the position of one object before.
The question format is presented as follows:

\vspace*{5pt}
\texttt{\textbf{Based on the agent's actions, does it know the position of <color> <name> before?}}
\vspace*{-5pt}
\begin{enumerate}[A., itemsep=-0.2em]
    \item \texttt{Yes}
    \item \texttt{No}
\end{enumerate}

The data for Task 8 are autonomously generated using seeds and a rule-based planner.

\subsubsection*{Task 9: Emotions}

This task (see Figure ~\ref{fig:task-emotion}) requires LLMs to infer the emotions of agents in a situated context from their physical behaviors. 
Specifically, it involves variations on a theme involving a small lake of water. 
In every variation, one agent pushes another agent into the lake. 
In some of these variations, an observer is privy to the situation. 
The prompt then asks the LLM about this observer's feelings towards both the victim and the perpetrator. 
Presumably, the observer should experience sympathy for the victim and anger (or a similarly negative emotion) for the perpetrator. 

\vspace*{5pt}
\texttt{\textbf{How would <observer.name> most likely feel about <pusher.name>?}}
\vspace*{-5pt}
\begin{enumerate}[A., itemsep=-0.2em]
    \item \texttt{no strong emotion}
    \item \texttt{angry}
\end{enumerate}

In other variations, a helper comes along and pulls the victim out of the lake. 
Presumably, the observer should feel positive emotions (e.g. respect, gratitude) for this helper.
The question format is as follows:

\vspace*{5pt}
\texttt{\textbf{How would <observer.name> most likely feel about <helper.name>?}}
\vspace*{-5pt}
\begin{enumerate}[A., itemsep=-0.2em]
    \item \texttt{no strong emotion}
    \item \texttt{respectful}
\end{enumerate}

\section{Prompting and Reproducibility}
\label{sec::prompting}

In this pilot study, our data curation follows a uniform structure across all tasks similar to prior work~\citep{li2022pre}, deviating only slightly to account for task-specific circumstances.\footnote{The data for this pilot study is available at \url{https://huggingface.co/datasets/sled-umich/2D-ATOMS}.}

\vspace*{-5pt}
\paragraph{Environment Description}
Each prompt begins with a description of the two-dimensional world wherein the task will take place. 
Specifically, our prompting code provides LLMs with the dimensions of the game board and a method to reference specific cells (column-first Cartesian coordinates). 
Each prompt subsequently itemizes the various objects that are situated in the world, along with their coordinates and attributes. 
Although this prompting structure could be easily adapted to handle many different types of attributes, we focused only on color for the sake of simplicity. 
Additionally, this section assigns each object a ``label'': a single letter that the prompt uses to represent the object in a printed grid representation.

\vspace*{-5pt}
\paragraph{Agent Description, Observability, and Task}
The next section of each prompt is a detailed description of the agent(s) occupying the grid world. 
In the example below, which was taken from task 1, only one agent occupies the grid world. 
In multi-agent tasks, this section details the various properties of all agents. Whether single or multi-agent, however, the basic properties are the same. 
The prompt first details the position and direction of the agent. 
It is located in a specific cell, and it always faces up, down, left, or right. The prompt then specifies the various actions that an agent is capable of taking (e.g. ``forward'', ``left'', and ``open''). 
The agent's labels are then specified. Within the printed grid representation, the agent is always represented as a V-like shape depending on the direction that it is facing. For instance, the prompting tool uses < to represent an agent that is facing left. 
Finally, the prompt specifies two key attributes of the agent: (1) its level of observability (e.g. whether or not it can see into adjacent rooms) and (2) its goal. 
Sometimes these two descriptors are heavily modified, restricted, or removed altogether so that they do not interfere with the task. 
For instance, when testing LLMs for percepts, the prompt does not specify the visual range of the agent.

\vspace*{-5pt}
\paragraph{Action Sequences}
Following the agent description section, the prompt prints out a board representation, a multi-line sequence of plain text that appears two-dimensional when printed out. 
Here, the various objects and agents are depicted in position by their associated labels. 
Additionally, the configuration of the walls is specified by a perimeter of W's.
Next, the prompt specifies a sequence of actions that take place over the course of the episode being considered by the LLM. 
In the episode depicted below, the agent navigates part of the way to a red box. 
These actions always specify the new position and orientation of the agent.

\vspace*{-5pt}
\paragraph{Questions and Answer Candidates}
Finally, each prompt contains a question that the LLM must answer. 
These questions are the most task-specific portion of the prompt, so their contents vary, however, they are always multiple-choice. 
Additionally, they always contain a set of instructions below heeding the LLM to return only a single letter in its response.

\clearpage

\begin{tcolorbox}[
    title={Sample Prompt for Task 1},
    halign=center,
    valign=center,
    nobeforeafter,
]
\begin{verbatim}
This is a grid-like 2D world
The grid world consists of 6 rows and 
6 columns, 0-based
We use (i,j) to represent the i-th 
column (from left to right) and j-th 
row (from top to bottom). 

The following is a list of objects in 
this world. Each line starts with the 
object's position and is followed by 
its attributes
(2, 3): key, grey; represented by this 
label: G
(4, 4): box, red; represented by this 
label: H

Walls are depicted using the symbol W

There is an agent at (2, 2) facing 
left

The agent can take the following 
actions:
- left: makes the agent face left of 
where it is currently facing
- right: makes the agent face right 
of where it is currently facing
- forward: makes the agent move one 
step in the direction it is currently 
facing
- open: makes the agent open a door 
that it is in front of
- pickup: makes the agent pick up 
the object that it is in front of
- drop: makes the agent drop an item 
that it is holding
- stay: makes the agent stay where it 
currently is for a timestep

The agent is represented by the 
following labels depending on which 
direction it is facing:
- Facing left: <
- Facing up: ^
- Facing right: >
- Facing down: v

\end{verbatim}
\end{tcolorbox}

\begin{tcolorbox}
\begin{verbatim}

The agent has full observability, 
meaning it can see the entire world

The agent has been instructed to 
navigate to one of the two objects 
in the environment, although you do 
not know which

This is the starting state of the 
board:
```
    0 1 2 3 4 5 
0 | W W W W W W 
1 | W O O O O W 
2 | W O < O O W 
3 | W O G O O W 
4 | W O O O H W 
5 | W W W W W W 
```
This list contains a sequence of 
actions taken by the agent
(Step 1) The agent took action left 
and is now at (2, 2) facing down
(Step 2) The agent took action left 
and is now at (2, 2) facing right
(Step 3) The agent took action forward 
and is now at (3, 2) facing right
(Step 4) The agent took action forward 
and is now at (4, 2) facing right
(Step 5) The agent took action right 
and is now at (4, 2) facing down

Which action will the agent take next? 
A: left 
B: right 
C: forward

Please ONLY respond using the letter 
corresponding to your answer
Do not generate any text other 
than the letter
\end{verbatim}
\end{tcolorbox}

\end{document}